\documentclass[runningheads]{llncs}

\usepackage[mobile]{eccv}

\usepackage{eccvabbrv}

\usepackage{graphicx}
\usepackage{booktabs}

\usepackage[accsupp]{axessibility}  

\usepackage[pagebackref,breaklinks,colorlinks,citecolor=eccvblue]{hyperref}

\usepackage{orcidlink}

\usepackage{graphicx}
\usepackage{booktabs}
\usepackage{multirow}
\usepackage{tikz}
\usepackage{pgfplots}
\usepackage{pgf}
\pgfplotsset{compat=newest}
\usepackage{pgfplotstable} 
\usepackage{color}
\usepackage{colortbl}

\definecolor{tabfirst}{rgb}{1, 0.7, 0.7} 
\definecolor{tabsecond}{rgb}{1, 0.85, 0.7} 
\definecolor{tabthird}{rgb}{1, 1, 0.7} 

\definecolor{custompurple}{HTML}{8D57CF}
\definecolor{customgreen}{HTML}{1E945B}
\definecolor{customred}{HTML}{DE333C}
\definecolor{customorange}{HTML}{DD752A}

\definecolor{plotred}{HTML}{e41a1c}
\definecolor{plotblue}{HTML}{377eb8}
\definecolor{plotgreen}{HTML}{4daf4a}
\definecolor{plotpurple}{HTML}{984ea3}

\begin{document}

\title{
E.T. the Exceptional Trajectories:\\
Text-to-camera-trajectory generation\\ with character awareness
}

\titlerunning{E.T. the Exceptional Trajectories}

\author{
Robin Courant\inst{1}
Nicolas Dufour\inst{1,2}
Xi Wang\inst{1}
Marc Christie\inst{3}
Vicky Kalogeiton\inst{1}
}

\authorrunning{R.~Courant et al.}

\institute{
LIX, Ecole Polytechnique, IP Paris \and
LIGM, Ecole des Ponts, CNRS, UGE \and 
Inria, IRISA, CNRS, Univ. Rennes
}

\maketitle
\begin{figure}
  \centering
  \vspace{-10pt}
    \includegraphics[width=\linewidth]{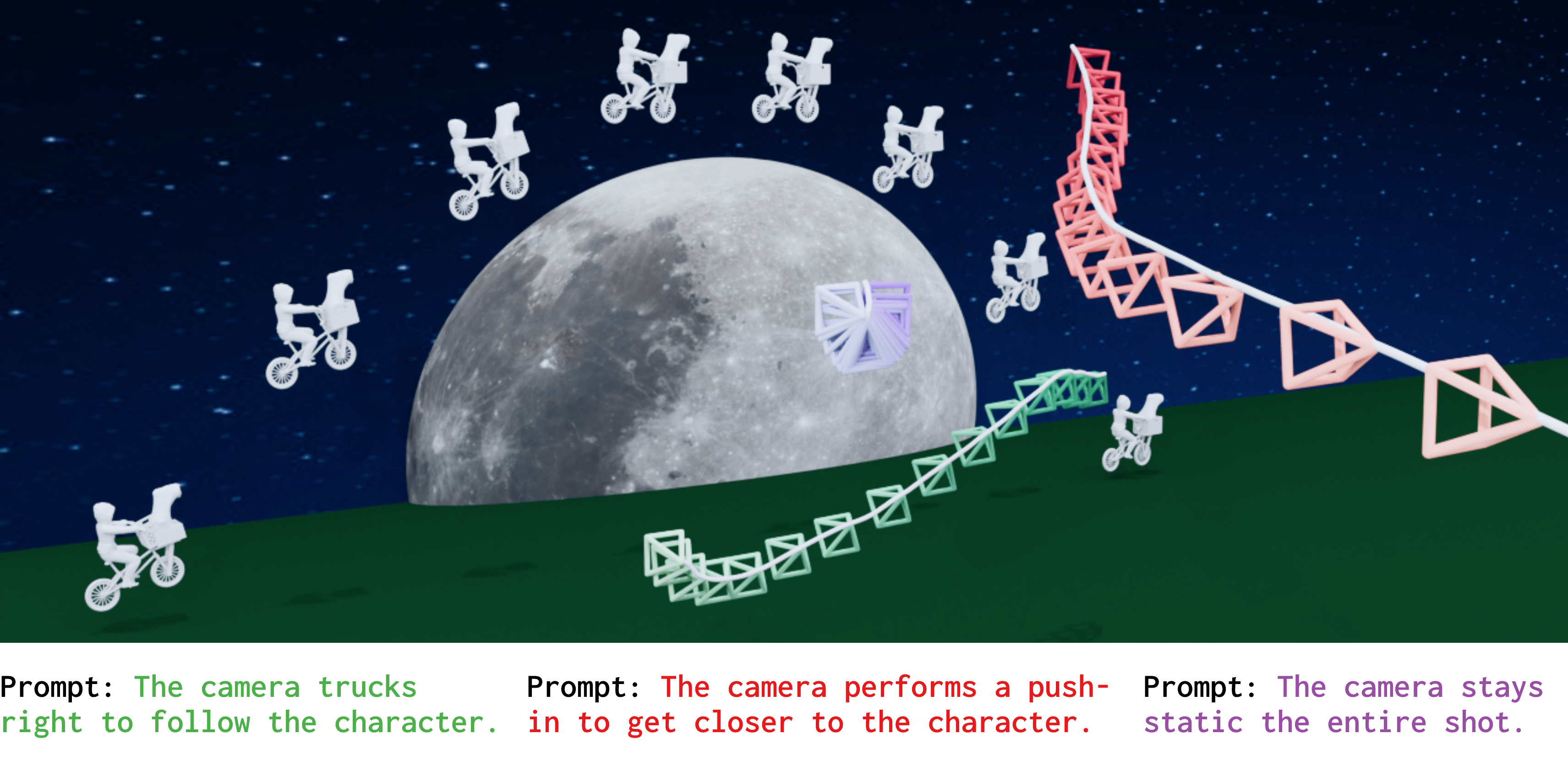}
    \caption{Different results generated by our camera trajectory diffusion system. 
    Project page \url{https://www.lix.polytechnique.fr/vista/projects/2024_et_courant}.
    }
    \label{fig:teaser}
    \vspace{-20pt}
\end{figure}

\begin{abstract}

Stories and emotions in movies emerge through the effect of well-thought-out directing decisions, in particular camera placement and movement over time. 
Crafting compelling camera trajectories remains a complex iterative process, even for skilful artists.
To tackle this, in this paper, we propose a dataset called the Exceptional Trajectories (E.T.) with camera trajectories along with character information and textual captions encompassing descriptions of both camera and character. 
To our knowledge, this is the first dataset of its kind. 
To show the potential applications of the E.T. dataset, we propose a diffusion-based approach, named {\sc Director}, which generates complex camera trajectories from textual captions that describe the relation and synchronisation between the camera and characters. 
To ensure robust and accurate evaluations, we train on the E.T. dataset CLaTr, a Contrastive Language-Trajectory embedding for evaluation metrics. We posit that our proposed dataset and method significantly advance the democratization of cinematography, making it more accessible to common users.

\end{abstract}


\section{Introduction}
\label{sec:introduction}
Cinematography is a collaborative and complex crafting process that mixes technical, artistic and storytelling skills. 
The ultimate objective is to communicate a distinct message to the audience, at a cognitive (e.g., revealing facts), emotional and aesthetic level, through tasks such as laying out the scene (\emph{mise-en-scène}), setting up the lighting and making decisions to place and move the camera in relation to the characters, their actions or the overall scene content.  
In this context, the camera is the only window into this staged world and therefore plays a critical role in conveying the director's intention. 
Through more than a hundred years of practice, cinematography has forged a common language for directors -- the \emph{film grammar} -- that prescribes how to place and move the camera to achieve intended effects. Yet mastering camera placements and motions remains challenging, especially for novice users confronted with hundreds of possibilities and little insights into how to generate the best ones. 

To lower the barriers in handling camera placement and camera motion, researchers have introduced a variety of methods. These include purely geometric approaches~\cite{blinn1988looking,lino2015intuitive}, optimization- and control-based strategies~\cite{drucker1992cinema, galvane2015camera}, as well as deep learning-grounded methodologies~\cite{jiang20sig,bonatti2020autonomous, Huang_2019_CVPR,drucker1992cinema} to interactively or automatically compute the parameters of camera trajectories. 
Typically, these methods address cinematographic tasks as either cinematic-rule-based control~\cite{Huang_2019_CVPR,bonatti2020autonomous,galvane2015camera} or example-based imitation~\cite{jiang20sig,jiang21siga,wang2023jaws}, conceptually resembling discriminative and regression models or registration and adaptation methods, respectively. Such techniques, however, suffer from the need to either design the underlying geometric model for each type of motion, or to design carefully crafted cost functions for each motion, and are often limited in their capacity to combine mixed motions creatively.

Recent advances in video generation~\cite{wang2023motionctrl,zhao2023motiondirector} enable users to explore more creative possibilities by capturing and reproducing camera motion in their generated videos. 
Jiang \emph{et al.} \cite{jiang2024ccd} followed this path and addressed camera trajectory generation using diffusion models, which incorporate a high degree of controllability. Yet, this work displayed two main drawbacks: first, it relied on a character-centric coordinate system to simplify the problem, thus limiting its generation capabilities, and second its evaluation metrics relied on camera trajectory features with oversimplified assumptions. 

In other domains, the generative techniques often rely on the availability of large datasets enriched with textual descriptions, such as language-motion obtained via motion capture (mocap)~\cite{plappert2016kit,guo2022humanml3d} or language-vision~\cite{lin2014mscoco,schuhmann2021laion} datasets. Yet in cinematography, there is no movie datasets where crucial cinematic information such as camera and character trajectories are available. Most recent approaches build on synthetic data \cite{jiang20sig,jiang21siga,jiang2024ccd}, or general videos from streaming platforms (see \cite{Huang_2019_CVPR} for drone trajectory generation, or \cite{zhou2018realestate10k} for dedicated real-estate videos) without the cinematic features that conform to the film grammar. Some example-based approaches address cinematic transfer tasks from real film clips~\cite{wang2023jaws,jiang2023cinematic}, these approaches only retarget and adapt the camera trajectory with little control or variability in the results and do not encode cinematographic knowledge.

\begin{figure}[t]
  \centering
    \includegraphics[width=\linewidth]{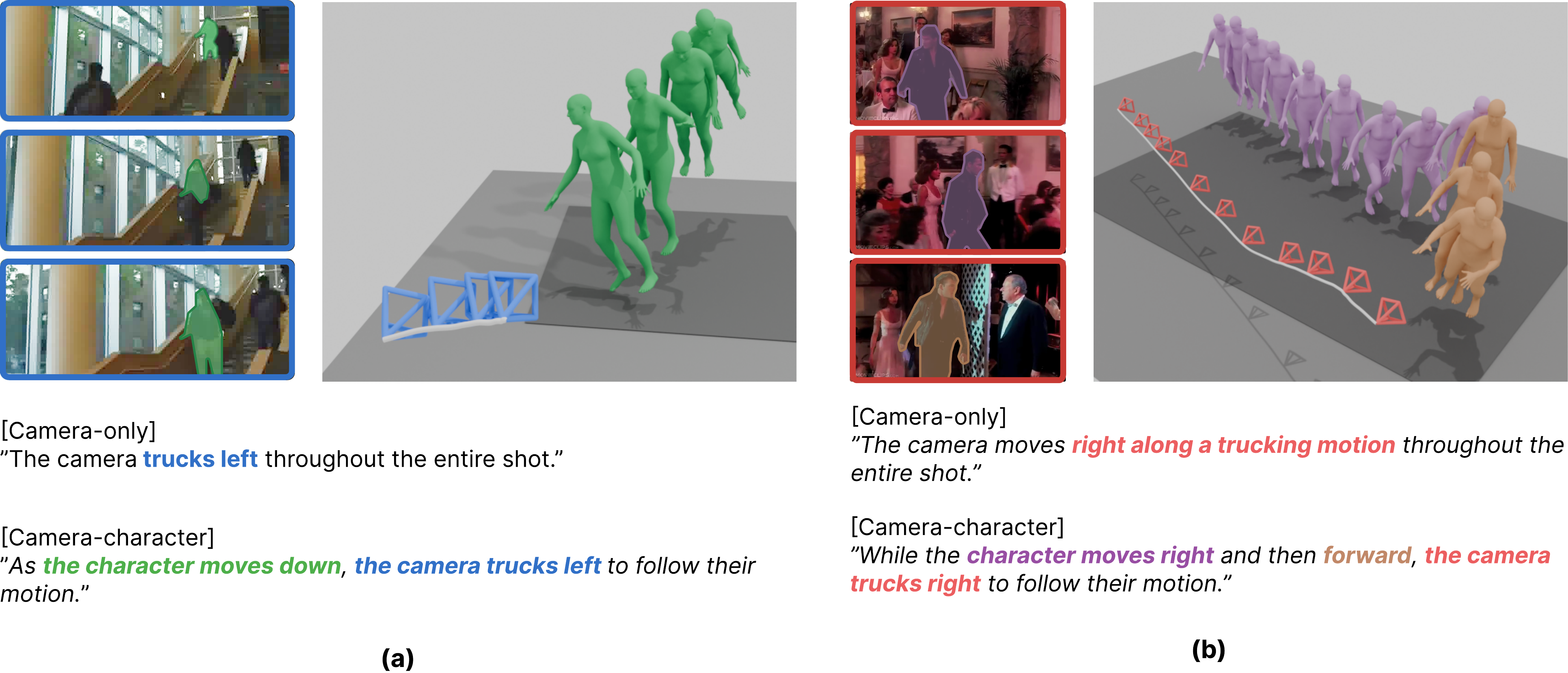}
    \caption{\textbf{Examples E.T. samples.} Each subfigure presents frames from the original movie shot on the left, while the right side depicts the extracted and processed camera and character trajectories. Additionally, the bottom part showcases the generated camera trajectory caption with or without the character trajectory.}
    \label{fig:example-dataset}
\end{figure}

In this work, we propose a new camera trajectory dataset extracted from real movie clips, called \emph{E.T. the Exceptional Trajectories}. It comprises camera trajectories together with textual descriptions of both camera and character trajectory over time (see Figure~\ref{fig:example-dataset}). E.T. contains more than $11$M frames with the corresponding camera and character trajectories, as well as two types of captions: camera-only and camera-character, describing the trajectory of the camera with respect to the trajectory of the character. To our knowledge, E.T. is the first extensive dataset with geometric information on both camera and character trajectories accompanied by textual descriptions.

To exploit this dataset, we also propose {\sc Director} (DiffusIon tRansformEr Camera TrajectORy), a diffusion-based model that generates camera trajectories by leveraging text descriptions and character information, as shown in Figure~\ref{fig:teaser}. 
This allows us to better encode the correlation between character and camera trajectories. 
Moreover, unlike previous methods~\cite{jiang2024ccd} that use a constrained character-relative coordinate system, we propose to use a global coordinate system. 
{\sc Director} relies on a classical diffusion framework with three distinct architectures for conditioning: in-context, AdaLN and cross-attention settings.  
Furthermore, we propose a language-trajectory embedding: CLaTr (Contrastive Language-Trajectory), trained at scale using the E.T. dataset. CLaTr serves as a foundation for computing default generative metrics similar to Frechet-Inception-Distance (FID)~\cite{heusel2017gans} for generated trajectories. 
Our experiments show that all three architectures of {\sc Director} successfully leverage the combination of input captions and character trajectories as conditions. Overall, {\sc Director} sets the new state-of-the-art on the camera trajectory generation task.

Our contributions are:
(1) We introduce the E.T. camera trajectory dataset extracted from real movie clips. 
We complement camera trajectories with character trajectories and captions for both camera and character.  
(2) We present {\sc Director}, a camera trajectory diffusion model that exploits both character trajectories and textual descriptions. It offers higher controllability and granularity for users than existing approaches~\cite{jiang2024ccd} and achieves state-of-the-art performances. 
(3) We propose CLaTr, a robust and accurate language-trajectory embedding, which facilitates the evaluation of camera trajectory generation models.

\section{Related work}
\label{sec:related-work}

\noindent \textbf{Camera control.} 
Over the past twenty years, there have been several paradigm shifts in camera planning and control.
Initial studies~\cite{blinn1988looking} predominantly focused on geometric modeling~\cite{lino2015intuitive} and rule-based trajectory controls~\cite{drucker1992cinema} to direct and create camera trajectories that comply with either hand-crafted cinematic rules or image-based criteria. 
With the progress of deep learning, \cite{jiang20sig} introduced a method to synthesize camera trajectory for 3D animations in two stages: (i) capturing cinematic styles from a reference clip using a Mixture-of-Experts model, and (ii) generating trajectories based on 3D character animations autoregressively. Subsequent research~\cite{jiang21siga}, building on this, incorporates keyframing to provide extra control such as positional and velocity constraints. More recently, JAWS~\cite{wang2023jaws} pioneered the direction for example-based camera retargeting within a Neural Radiance Field (NeRF)~\cite{mildenhall2020nerf} setting, by optimising camera trajectory directly given the 2D reference clip in a 3D NeRF. All these example-based methods share a common limitation: they struggle with generalization because they require carefully selected reference videos to ensure high quality. 

Unlike example-based methods, many cinematic-rule-based methods readily integrate with Deep Reinforcement Learning (DRL) and Imitation Learning (IL) techniques, particularly in the drone cinematography domain: \cite{Huang_2019_CVPR} exploit optical flow and human poses to guide drone controls via an IL framework. Similarly, \cite{bonatti2020autonomous} use DRL to control drone actions for multiple rewards, including obstacle avoidance, target tracking, shooting style etc. Recently, GAIT~\cite{Xie_2023_ICCV} employs an aesthetic score-based RL method instead of handcrafted rewards to control the camera in the virtual 3D environment. 
However, these RL-based camera control approaches also have limitations: (1) they need environment-specific training; 
(2) they inherently restrict the diversity of results, often leading to collapsed trajectory styles.
Instead, we leverage the generalization capabilities of generative models to address the camera control task.

\noindent \textbf{Camera diffusion.}
Generative models have recently gained much progress and attention in domains such as textual-conditioned image generation~\cite{rombach2022ldm,podell2023sdxl,nichol2022glide}, video synthesis~\cite{singer2022make,blattmann2023stable} and human motion generation~\cite{tevet2022mdm,chen2023executing,zhang2024modi}. Among these, diffusion models stand out for their strong ability to produce high-fidelity and diverse generative samples~\cite{xiao2021tackling,dhariwal2021diffusion}, making them particularly well-suited for camera trajectory generation tasks. 

The first application of diffusion models in camera control is the Cinematographic Camera Diffusion (CCD)~\cite{jiang2024ccd}, which relies on the MDM architecture (human Motion Diffusion Model)~\cite{tevet2022mdm} and is trained on synthetic data. However, CCD simplifies the task by expressing all the camera trajectories in character-centric relative coordinates. 
Its small-scale synthetic training dataset also limits the broader application of the method (e.g., only 48-size vocabulary is used during training), thus making it unable to generate camera trajectories from real datasets and, in turn, impractical for common users. 
In contrast, in our proposed E.T. dataset, we represent 
camera trajectories in a global coordinate system, distinct from character trajectories. This approach allows for more diverse correlations between character and camera movements. Additionally, E.T. offers a rich vocabulary ($\sim$ 5.4k) and extensive camera trajectory data.

Recent literature also includes several text-to-video generation techniques that can handle different categories of camera motions~\cite{wang2023motionctrl,zhao2023motiondirector}. These methods, however, assume access to 3D camera trajectories, whereas our approach generates them. 
Furthermore, they typically overlook the camera's primary targets (i.e., the characters), which are essential for defining camera trajectories. 
In contrast, our dataset contains character information, and we leverage it to generate camera trajectories that focus on a specific target character.

\noindent \textbf{Camera trajectory datasets.}
Many modern generative methods leverage large multimodal datasets. 
For instance, in text-to-image generation, the default dataset is LAION~\cite{schuhmann2021laion} with around 400 million image-text pairs. Similarly, in human motion synthesis, the large-scale KIT~\cite{plappert2016kit} and HumanML3D~\cite{guo2022humanml3d} datasets offer detailed textual captions that enhance comprehension of human motion. Yet, for camera control, only a few datasets are available~\cite{zhou2018realestate10k,jiang2024ccd}. This is largely due to the intricacies involved in extracting camera poses from real-world videos, especially in cinematic contexts due to the presence of stylistic elements (\eg motion blur or depth-of-field). 
Zhou et al.~\cite{zhou2018realestate10k} applied Structure-from-Motion (SfM) methods to YouTube real-estate videos, creating the RealEstate10K dataset. This dataset, designed primarily for 3D reconstruction, comprises solely smooth camera movements and limited scene variation, lacking the nuanced complexity of cinematic camera motion and human presence.
More recently Jiang et al.~\cite{jiang2024ccd} introduced a synthetic cinematic camera trajectory dataset, aiming to circumvent extraction challenges. However, this dataset oversimplifies the intricate cinematic dynamics present in real-world movies.

A recent breakthrough in 3D human pose estimation for videos, termed SLAHMR~\cite{goel2023humans4d}, offers a compelling trade-off between robustness and accuracy by jointly optimizing camera and character trajectory estimations.  
Motivated by the lack of camera trajectory datasets, the capabilities of SLAHMR and the recent advances in other domains, we propose a new multi-modal camera trajectory dataset E.T. extracted from cinematic content, which we enhance with automatically generated captions for camera and character trajectories.

\section{Exceptional Trajectories (E.T.)}
\label{sec:dataset}
\begin{table}[t]
\centering
\resizebox{\textwidth}{!}{%
    \begin{tabular}{lcccc@{\hspace{10pt}}c@{\hspace{6pt}}c@{\hspace{10pt}}c@{\hspace{6pt}}c@{\hspace{10pt}}c}
    \toprule
    \multirow{2}{*}{\textbf{Dataset}} & \multirow{2}{*}{\textbf{\#Samples}} & \multirow{2}{*}{\textbf{\#Frames}} & \multirow{2}{*}{\textbf{\#Hours}} & \multirow{2}{*}{\textbf{Domain}} & \multicolumn{2}{c}{\textbf{Character}} & \multicolumn{2}{c}{\textbf{Camera}} & \multirow{2}{*}{\textbf{\#Vocabulary}} \\
          &           &          &         &        & Traj  & \#Captions & Traj & \#Captions & \\
    \midrule
    KIT Motion-Language~\cite{plappert2016kit} & 4K & 0.8M & 11.23 & Mocap & \checkmark & 6K & & - & 1,623 \\
    HumanML3D~\cite{guo2022humanml3d} & 14K & 2M & 28.59 & Mocap & \checkmark & 45K & & - & \textbf{5,371} \\
    RealEstate10k~\cite{zhou2018realestate10k} & 79K & \textbf{11M} & \textbf{121} & Youtube & & - & \checkmark & - & - \\
    CCD~\cite{jiang2024ccd} & 25K & 4.5M & 50 & Synthetic & & - & \checkmark & 25K & 48 \\
    \midrule
    E.T. (Ours) & \textbf{115K} & \textbf{11M} & 120 & \textbf{Movie} & \checkmark & \textbf{115K} & \checkmark & \textbf{230K} & 1,790 \\
    \bottomrule
    \end{tabular}%
}
\caption{\textbf{Dataset comparison.}
We compare the E.T. dataset to (i) two human motion datasets KIT~\cite{plappert2016kit} and HumanML3D~\cite{guo2022humanml3d}; and (ii) camera trajectory datasets RealEstate10K~\cite{zhou2018realestate10k} and CCD~\cite{jiang2024ccd}. Here the notion of sample is common across all datasets and corresponds to data associated with a continuous temporal sequence.
}
\label{tab:datasets}
\end{table}

We introduce a camera trajectory dataset called \emph{Exceptional Trajectories} (E.T.), extracted from real movies. E.T. is built upon the Condensed Movies Dataset (CMD)~\cite{bain2020cmd}. Each \emph{sample} in E.T. represents a camera trajectory at the shot level together with a character trajectory and two types of textual captions: a camera-only caption, which describes the camera motion; and a joint camera-character trajectory caption, which describes the motion of the camera according to the motion of the character (see Figure~\ref{fig:example-dataset}). Below, we describe the key properties and statistics of E.T. (Section~\ref{sub:properties}) followed by the creation pipeline (Section~\ref{sub:dataset-creation}). 

\subsection{E.T. properties and statistics}
\label{sub:properties}
The key properties of E.T. are as follows:

\noindent \textbf{Cinematic content.} 
The camera trajectories in E.T. are both realistic and cinematic, since they are extracted from real-world movies (Table~\ref{tab:datasets}). This dual nature allows for effective modelling of various visual styles, in contrast to RealEstate10k's~\cite{zhou2018realestate10k} focus on shots characterized by smooth camera trajectories and limited scene variation. 
Furthermore, by extracting data from real-world movies, E.T. sets itself apart from CCD~\cite{jiang2024ccd}, which only relies on synthetic camera trajectories.

\noindent \textbf{Scale.}
E.T. is built upon $16,210$ different scenes from CMD~\cite{bain2020cmd}. 
It comprises $115$K samples spanning $11$M frames and totalling $120$ hours of footage, offering extensive and diverse camera and character (human) trajectories based on real movies. In contrast, existing human motion datasets are much smaller, with only $11.23$ hours for KIT~\cite{plappert2016kit} and $28.59$ hours for HumanMl3D~\cite{guo2022humanml3d} (see Table~\ref{tab:datasets}). When compared against datasets with camera trajectories, it far exceeds CCD~\cite{jiang2024ccd} in terms of hours, frames and samples. Although its scale is comparable to RealEstate10k~\cite{zhou2018realestate10k}, it provides additional character trajectories and captions referring to real movies as opposed to RealEstate10k, which focuses only on camera trajectories in another domain. 

\noindent \textbf{Controllability.}
E.T. stands out by comprising not only camera and character trajectories but also camera-only and camera-character captions (see Figure~\ref{fig:example-dataset}).
Incorporating caption information into the model offers multiple advantages: (1) it democratizes the input format for general users; and (2) it adds complementary semantic information to the trajectory data.
In comparison, RealEstate lacks captions entirely. CCD's captions are limited by a small vocabulary size and focus only on camera while lacking character information\footnote{Note that CCD indirectly comprises camera trajectories through the character-relative coordinate system.}. 
The richness and complexity of E.T.'s captions are on par in terms of vocabulary size --above a thousand-- with human motion datasets such as KIT and HumanML3D, which provide detailed, hand-crafted human motion descriptions\footnote{Note that E.T. has no overlap with human motion datasets. E.T.'s extracted 3D poses (see Section~\ref{sub:dataset-creation}) are less accurate than the ones in motion capture, while its captions describe camera trajectory relative to character trajectory, as opposed to describing exact human motions targeted by these datasets.}.

\noindent \textbf{Statistics.} 
Figures~\ref{fig:cam-segments}-~\ref{fig:character-segments} display the statistics of the E.T. dataset, confirming the diversity and all six degrees of freedom coverage of both camera and character trajectories (see more in Appendix~\ref{supp:sub:stats}.)

\subsection{Dataset creation pipeline}
\label{sub:dataset-creation}

\begin{figure}[t]
  \centering
    \includegraphics[width=\linewidth]{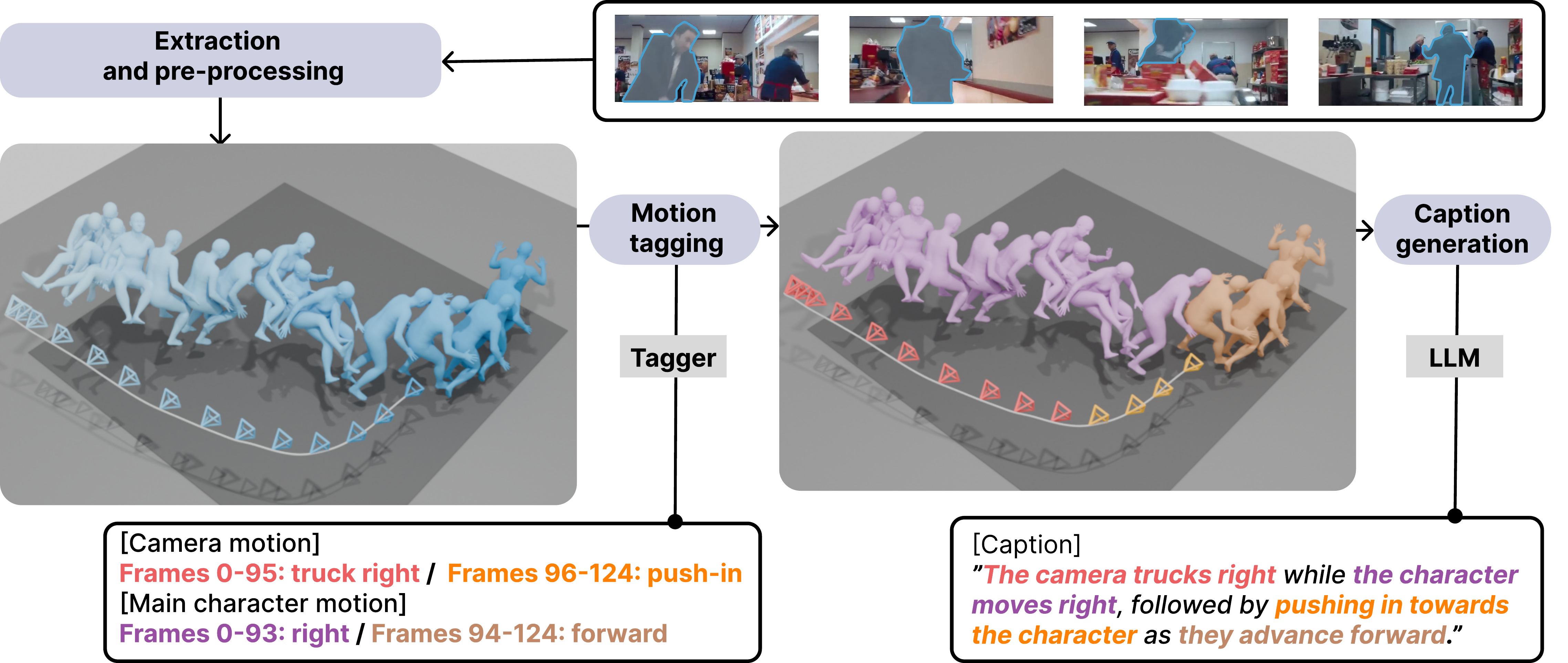}
    \caption{\textbf{Dataset creation pipeline.} Given RGB frames from a video, we first extract and pre-process camera and character poses, then tag resulting camera and character trajectories (sequence of poses) to obtain rough independent descriptions (middle part). Finally, we translate these descriptions into rich textual captions, aligning the camera trajectory with that of the character (right part). }
    \label{fig:main-dataset}
\end{figure}

E.T. is constructed by a three-step process (see Figure~\ref{fig:main-dataset}). First, we extract the 3D coordinates of cameras and characters over time, which we further refine to form uniform trajectories. Second, we perform \emph{motion tagging}, i.e. partition each trajectory into segments with each segment comprising a pure camera motion that we label (tag).  
Third, we generate captions that describe both the camera and the character trajectory over time.  
We detail each step below. 

\paragraph{Data extraction and pre-processing.} 
To extract camera and character poses, we apply on each shot the joint camera and 3D human poses estimator SLAHMR~\cite{ye2023slahmr}.
Given the complexity of estimating 3D poses from 2D data, the raw outputs tend to be noisy. To address this, we perform various pre-processing steps such as alignment, filtering, smoothing and cropping to a maximum length of 300 frames as in~\cite{guo2022humanml3d}. Refer to the Appendix~\ref{supp:sub:processing} for further details.

\begin{figure}[t]
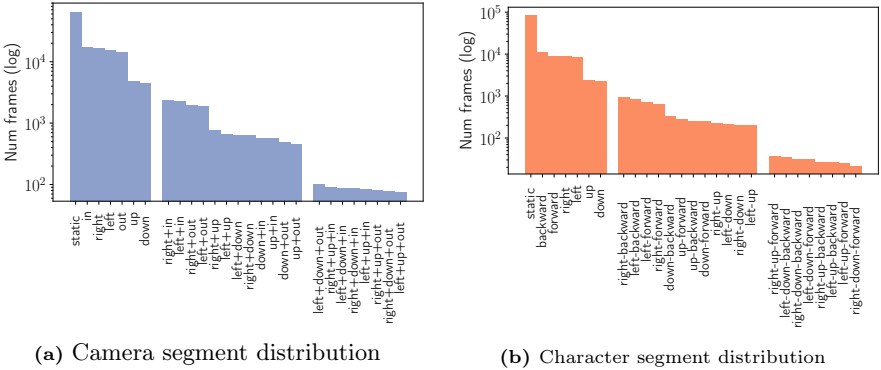

\centering

\begin{subfigure}[b]{.49\linewidth}
\scalebox{.37}{\input{fig/cam_segments_hist.pgf}}
\caption{\small{Camera segment distribution}}\label{fig:cam-segments}
\end{subfigure}
\begin{subfigure}[b]{.49\linewidth}
\scalebox{.37}{\input{fig/char_segments_hist.pgf}}
\caption{Character segment distribution}\label{fig:character-segments}
\end{subfigure}

\caption{\textbf{E.T. statistics.} 
}
\label{fig:dataset-distribution}
\end{figure}

\paragraph{Motion tagging.}
Our objective is to partition camera or character trajectories into segments of pure motion: tags. Besides static, we consider the six fundamental motions across three degrees of freedom. They include lateral movements left, and right; vertical movements up and down; and depth movements forward and backwards. Each trajectory is partitioned into motion tags with one, two, or three pure camera motions, totalling 27 combinations (see Figure~\ref{fig:cam-segments}). 

We propose a thresholding-based method that uses trajectory velocity for motion tagging: This method consists of two stages: (i) for each dimension (XYZ), we use an initial threshold on velocity to detect whether the camera or character remains static along the dimension; (ii) when multiple dimensions are non-static, we calculate pairwise velocity rates and use a threshold to pinpoint dominant velocities. A dimension is classified as static if its velocity is outmatched. 
The tag of motion between two points is then determined by the combination of non-static dimensions.
Finally, we apply smoothing to avoid noisy and sparse tags and hence enhance the overall trajectory-level tagging. 

For \textit{camera trajectory tagging}, we use the rigid body velocity $\in {SE(3)}$ -- derived from rotation and translation-- to account for the camera's facing direction. 
this enables us to differentiate between similar motions, such as `trucking', where the camera moves along an axis with a perpendicular facing direction, and `depth', where the facing direction aligns with the movement axis. 
For \textit{character trajectory tagging}, we assume that characters face the direction of their movement. Hence, we represent character trajectory using only the linear velocity, as derived from the translation of their hip centres. 

These result in a coarse description of both camera and character trajectories over time as shown in Figure~\ref{fig:main-dataset} (left). 

\paragraph{Caption generation.} 
Our objective is to provide rich textual descriptions of the extracted camera trajectories according to the character trajectory. 
In movie, cameras typically move relative to the subject being filmed, i.e., the main character. 
Therefore, for each shot, we first identify the main character following~\cite{truffaut1985hitchcock}\footnote{Hitchcock's rule: `\textit{the size of an object in the frame should equal its importance in the story at the moment}'~\cite{truffaut1985hitchcock}.} based on the temporal and spatial coverage of their bounding boxes within the shot. 
Then, for both camera and main character trajectories, we generate captions for each motion tag, as shown in the center of Figure~\ref{fig:main-dataset}. 
Then, inspired by~\cite{delmas2022posescript}, our goal is to convert the descriptions obtained via motion tagging for camera and character trajectories into detailed textual annotations. For this, we prompt an LLM --Mistral-7B~\cite{jiang2023mistral}-- to generate camera trajectory captions by referencing the main character's trajectory as anchor points. 
Our prompt formulation follows a structured approach with context, instruction, constraint, and example. Further details can be found in the Appendix~\ref{supp:sub:pipeline}. 

This step results in a rich description of both camera and character trajectories over time as shown in Figure~\ref{fig:main-dataset} (right). 

\section{Method}
\label{sec:method}
Here, we introduce our proposed \emph{DiffusIon tRansformEr Camera TrajectORy} ({\sc Director}) method for camera trajectory generation (Section~\ref{sub:camera_diffusion}). {\sc Director} takes as input the character trajectory with the camera-character caption and generates a camera trajectory. Additionally, we present the \emph{Contrastive Language-Trajectory} embedding (CLaTR) that serves as a basis for creating a common space between text and trajectories (Section~\ref{sec:clatr}), enabling the computation of evaluation metrics.

\subsection{Camera trajectory diffusion}
\label{sub:camera_diffusion}
\paragraph{Problem formulation.}
We consider a camera trajectory $\mathbf{x}_{1:N}$ as a sequence of $N$ consecutive camera poses. Each camera pose $\mathbf{x} = [\mathbf{R}|\mathbf{t}]$ comprises a rotation $\mathbf{R}$ representing the camera's orientation and a translation $\mathbf{t}$ indicating its position. We aim at generating camera trajectories under two conditions: (i) a target character trajectory $\mathbf{h}_{1:N}$ capturing the 3D positions of the main character; and (ii) a textual description $c$ specifying the desired camera movement relative to the character movement.

\paragraph{Diffusion framework.}
We follow the general diffusion paradigm established in EDM~\cite{karras2022edm}.
In essence, diffusion models consist of randomly sampling $\mathbf{x}^0 \sim \mathcal{N}(\mathbf{0}, \sigma_{max}^2 \mathbf{I})$,
and progressively denoising it to reach the endpoint $\mathbf{x}^K$ of this process, distributed according to the initial data distribution.
During the training stage, we perturb an initial data distribution with standard deviation $\sigma_{\text{data}}$, with i.i.d. Gaussian noise with standard deviation $\sigma$. When $\sigma_{\text{max}} \gg \sigma_{\text{data}}$, the noise distribution equivalent to a normal distribution $\mathcal{N}(\mathbf{0}, \sigma_{max}^2 \mathbf{I})$. 
We use these modified versions of the initial data distribution to train a denoiser module $D$, which takes as input a sample $\mathbf{x}$ to denoise, the two conditions (character trajectory $\mathbf{h}$ and the caption $c$), and the corresponding standard deviation $\sigma$. Then, $D$ is trained using the denoising score matching loss:
\begin{equation}
  \mathcal{L}_{\text{score}} = \big( D(\mathbf{x}, \mathbf{h}, c; ~\sigma) - \mathbf{x} \big) / \sigma^2
  \text{.}
\end{equation}

During the sampling phase, we apply the 2nd order deterministic sampling introduced in EDM~\cite{karras2022edm} with classifier-free guidance~\cite{ho2021cfg}.

\begin{figure}[t]
    \centering
    \begin{subfigure}[b]{0.32\textwidth}
        \includegraphics[width=\textwidth]{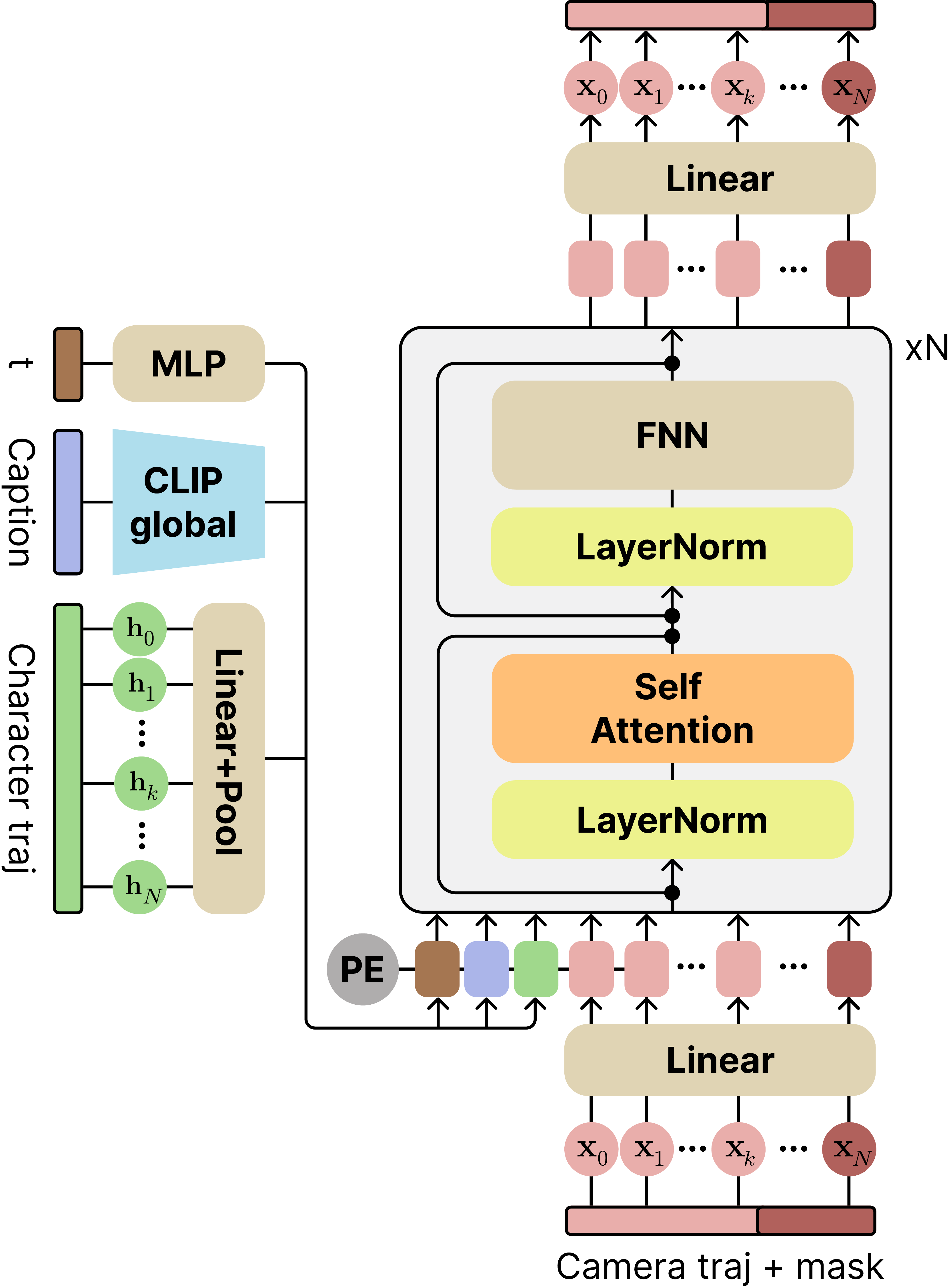}
        \caption{{\sc Director} A}
        \label{fig:incontext_director}
    \end{subfigure}
    \hfill
    \begin{subfigure}[b]{0.32\textwidth}
        \includegraphics[width=\textwidth]{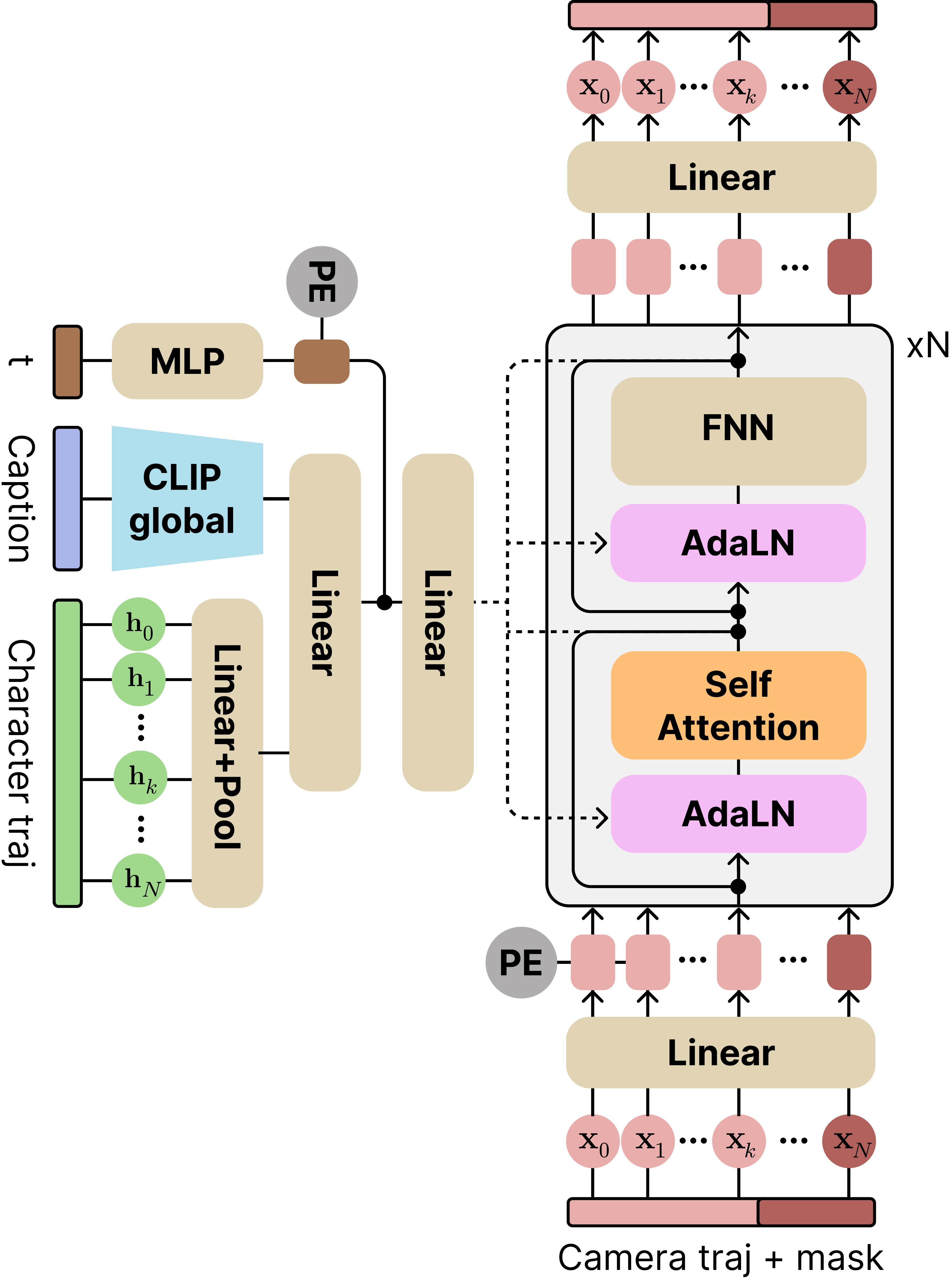}
        \caption{{\sc Director} B}
        \label{fig:adaln_director}
    \end{subfigure}
    \hfill
    \begin{subfigure}[b]{0.32\textwidth}
        \includegraphics[width=\textwidth]{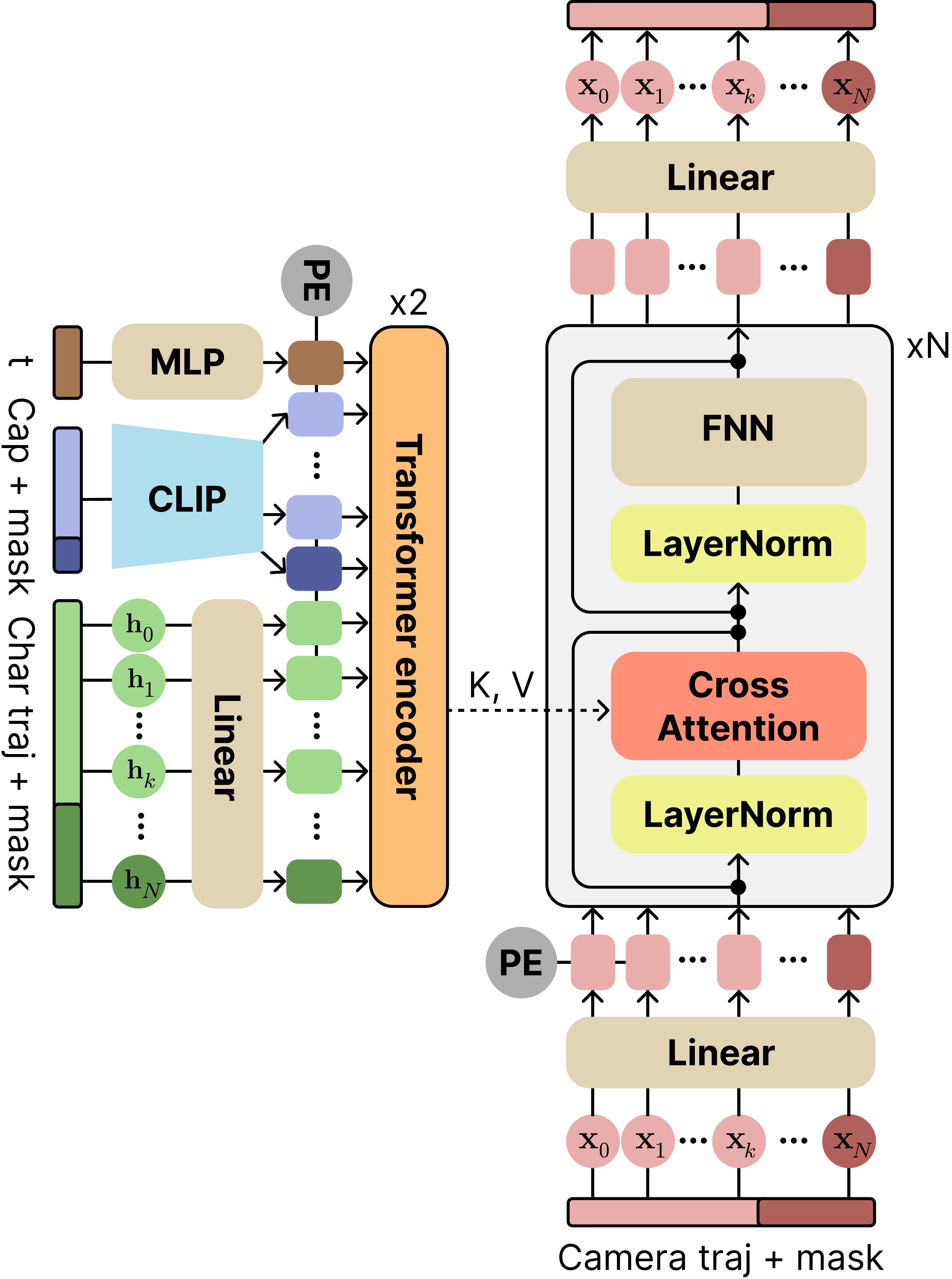}
        \caption{{\sc Director} C}
        \label{fig:cross_attention_director}
    \end{subfigure}
    \caption{\textbf{\emph{DiffusIon tRansformEr Camera TrajectORy} ({\sc Director}).} We display 3 variants of our diffusion model {\sc Director}. {\sc Director} A incorporates the conditioning as in-context tokens. {\sc Director} B leverages AdaLN modulation of the transformer block to add the conditioning. {\sc Director} C uses the full text and character trajectory sequences by relying on cross-attention. 
    }
    \label{fig:director}
\end{figure}

\paragraph{{\scshape Director} architecture.} 
{\sc Director} (\emph{DiffusIon tRansformEr Camera TrajectORy}) takes as input the character trajectory and the caption and generates a camera trajectory. Its architecture is illustrated in Figure~\ref{fig:director}. 
The base of {\sc Director} is a pre-norm Transformer~\cite{vaswani2017attention, xiong2020layer}. We condition the transformer on the diffusion timestep, the character trajectory, and a textual description that describes the relative movement between the camera and character trajectories (see Figure~\ref{fig:example-dataset}). The timestep is tokenized using a sinusoidal positional embedding~\cite{vaswani2017attention} and then mapped with an MLP. 

Inspired by the DiT architecture variants~\cite{peebles2023scalable}, we explore three distinct ways to include the conditioning in the denoising process (Figure~\ref{fig:director}). 

\noindent \textbf{{\scshape Director} A (Figure~\ref{fig:incontext_director}).} The conditioning is added to the \emph{\textbf{context}} of the transformer input. We only use the global clip token for the text, and we do a linear embedding of the character trajectories, which in turn gets averaged pooled into a single token. 

\noindent \textbf{{\scshape Director} B (Figure~\ref{fig:adaln_director}).} 
Both conditionings (character trajectory and caption) are concatenated into a single token which gets mapped at each layer into 6 vectors, $\gamma_1, \beta_1,\ \lambda_1, \gamma_2, \beta_2,\ \lambda_2$. Then, the layer-norm of the transformer is replaced by the following \emph{\textbf{AdaLN}} operation:  
\begin{equation}
    \text{ADALN}(\gamma, \beta, x) = (1+\gamma) \text{LN}(X)  + \beta \quad , 
\end{equation}

\noindent where LN refers to the Layer Normalization, 
$\gamma, \beta$ are the scale and bias, respectively. 
The AdaLN operation is performed before each self-attention and feed-forward layer in the transformer. The output of each self-attention and cross-attention is rescaled by $\lambda$. Following~\cite{peebles2023scalable}, we initialize the modulation such that the output is zero.

\noindent \textbf{{\scshape Director} C (Figure~\ref{fig:cross_attention_director}).}
We leverage the full sequence length of the conditioning. We retrieve the CLIP-embedded text sequence and the linearly projected trajectory and concatenate them into a single sequence. We then use 2 layers of transformer encoders to pre-process this sequence, which is then incorporated into the {\sc Director} transformer with a \emph{\textbf{cross-attention}} block.

\subsection{Contrastive Language-Trajectory embedding (CLaTr)}
\label{sec:clatr}

Given the scarcity of relevant camera trajectory methods and datasets, the community has not introduced adequate metrics for this task. 
In the concurrent cinematic camera trajectory diffusion work~\cite{jiang2024ccd}, the authors evaluate their model with metrics from the human motion community. For this, they train a dedicated camera trajectory classifier to extract features. However, their classifier is trained on a simplistic task, comprising only six basic camera motion classes on synthetic data, which fails to capture the true complexity of camera trajectories.

To address this lack of proper evaluation metrics, in this section, we propose to extend existing metrics from text-image-based and text-motion-based generation (which rely on feature embeddings to measure the generation quality) to text-trajectory generation. The main obstacle is that no commonly accepted text-trajectory feature embedding exists. Therefore, we propose to learn a general text-trajectory embedding in a contrastive CLIP-like manner to acquire an accurate and robust feature representation, which can serve as a foundation for computing camera trajectory evaluation metrics. 

We introduce \textit{Contrastive Language-Trajectory} embedding (CLaTr) by capitalizing our multi-modal dataset E.T. with a CLIP-like approach~\cite{radford2021clip}.  
Our language-trajectory embedding follows the methodology outlined in~\cite{petrovich2023tmr}, originally designed for human motion. 
CLaTr consists of a VAE~\cite{kingma2014auto} framework with trajectory and text encoders and a shared feature decoder.
CLaTr is trained with three losses: (a) a reconstruction loss $\mathcal{L}_R$, quantifying trajectory reconstruction of both trajectory and text features; (b) four KL loss terms $\mathcal{L}_{KL}$, which regularize each modality distribution and also enforce inter-modality similarity; and (c) a cross-modal embedding similarity loss $\mathcal{L}_E$, ensuring alignment between text and trajectory features. See Appendix~\ref{sup:sub:clatr} for more details.

\section{Experiments}
\label{sec:experiments}
\newcolumntype{A}{>{\hspace{6pt}}c<{\hspace{6pt}}}
\newcolumntype{B}{>{\hspace{3pt}}c<{\hspace{3pt}}}
\newcolumntype{C}{>{\hspace{3pt}}c<{\hspace{3pt}}}
\newcolumntype{S}{>{\hspace{0pt}}c<{\hspace{3pt}}}

\begin{figure}[htbp]
  \begin{minipage}[b]{.65\linewidth}
    \centering
    \resizebox{\linewidth}{!}{
    \begin{tabular}{@{}Sl|@{\hspace{4pt}}c@{\hspace{4pt}}|cAAAA|ABBc}
    \toprule
         \multirow{2}{*}{\textbf{Set}} & \multirow{2}{*}{\textbf{Methods}} & \multirow{2}{*}{$\mathbf{\omega}$} & \multicolumn{5}{c|}{\textbf{Camera trajectory quality}} & \multicolumn{4}{c}{\textbf{Text-camera coherence}} \\
         &  &  & FD\textsubscript{CLaTr} $\downarrow$    & P $\uparrow$ & R $\uparrow$     & D $\uparrow$     & C $\uparrow$     & CS $\uparrow$     & C-P $\uparrow$     & C-R $\uparrow$     & C-F1 $\uparrow$ \\ \hline
        \multirow{5}{*}{\rotatebox{90}{\textbf{\shortstack{E.T. pure\\trajectories}}}} 
          & CCD~\cite{jiang2024ccd}   & 5.5 & 31.33 & 0.79 & 0.55 & 0.83 & 0.72 & 3.21 & 0.53 & 0.28 & 0.27 \\
        ~ & MDM~\cite{tevet2022mdm}  & 1.8 &   6.10 &   0.77 & \cellcolor{tabsecond}0.68 &  0.89 &                      0.80 &                      21.26 &                      0.81 &  0.75 &  0.76 \\
        ~ & {\scshape Director} A  & 1.6 & \cellcolor{tabsecond}5.16 & \cellcolor{tabsecond}0.82 &   0.67 &  \cellcolor{tabfirst}1.00 & \cellcolor{tabsecond}0.86 & \cellcolor{tabsecond}21.88 & \cellcolor{tabsecond}0.84 & \cellcolor{tabsecond}0.78 & \cellcolor{tabsecond}0.80 \\
        ~ & {\scshape Director} B  & 1.8 &                      6.61 &   0.80 &  \cellcolor{tabfirst}0.72 & \cellcolor{tabsecond}0.92 &   0.82 &  \cellcolor{tabfirst}23.10 &  \cellcolor{tabfirst}0.85 &  \cellcolor{tabfirst}0.80 &  \cellcolor{tabfirst}0.86 \\
        ~ & {\scshape Director} C  & 1.6 &  \cellcolor{tabfirst}4.57 &  \cellcolor{tabfirst}0.83 &                      0.65 &  \cellcolor{tabfirst}1.00 &  \cellcolor{tabfirst}0.87 &   21.49 &   0.83 & \cellcolor{tabsecond}0.78 & \cellcolor{tabsecond}0.80 \\ \hline
        \multirow{5}{*}{\rotatebox{90}{\textbf{\shortstack{E.T. mixed\\trajectories}}}} 
          & CCD~\cite{jiang2024ccd}  & 6.0 & 35.81 & 0.73 & 0.55 & 0.75 & 0.67 & 6.26 & 0.37 & 0.20 & 0.17 \\
        ~ & MDM~\cite{tevet2022mdm}      & 2.0 &                      6.79 &   0.78 &                      0.65 &   0.85 &                      0.76 &                      18.32 &                      0.36 &                      0.36 &                      0.34 \\
        ~ & {\scshape Director} A  & 1.4 & \cellcolor{tabsecond}3.88 & \cellcolor{tabsecond}0.82 & \cellcolor{tabsecond}0.68 & \cellcolor{tabsecond}0.98 & \cellcolor{tabsecond}0.85 &   20.76 & \cellcolor{tabsecond}0.43 & \cellcolor{tabsecond}0.43 & \cellcolor{tabsecond}0.42 \\
        ~ & {\scshape Director} B  & 1.6 &   6.10 &   0.78 &  \cellcolor{tabfirst}0.74 &   0.85 &   0.78 & \cellcolor{tabsecond}20.78 &   0.41 &   0.40 &   0.39 \\
        ~ & {\scshape Director} C  & 1.4 &  \cellcolor{tabfirst}3.76 &  \cellcolor{tabfirst}0.83 &   0.67 &  \cellcolor{tabfirst}1.00 &  \cellcolor{tabfirst}0.86 &  \cellcolor{tabfirst}21.95 &  \cellcolor{tabfirst}0.49 &  \cellcolor{tabfirst}0.49 &  \cellcolor{tabfirst}0.48 \\
        \bottomrule
    \end{tabular}}
    \vspace{4pt}
    \captionof{table}{
    \textbf{Quantitative Results.} Comparison of {\scshape Director} and concurrent methods on E.T. pure and mixed subsets, evaluating trajectory quality (left) and caption coherence (right). \colorbox{tabfirst}{First best} and \colorbox{tabsecond}{second best}.}
    \label{tab:quant}
  \end{minipage}
  \hfill
  \centering
  \begin{minipage}[b]{.33\linewidth}
    \centering
    \includegraphics[width=0.99\linewidth]{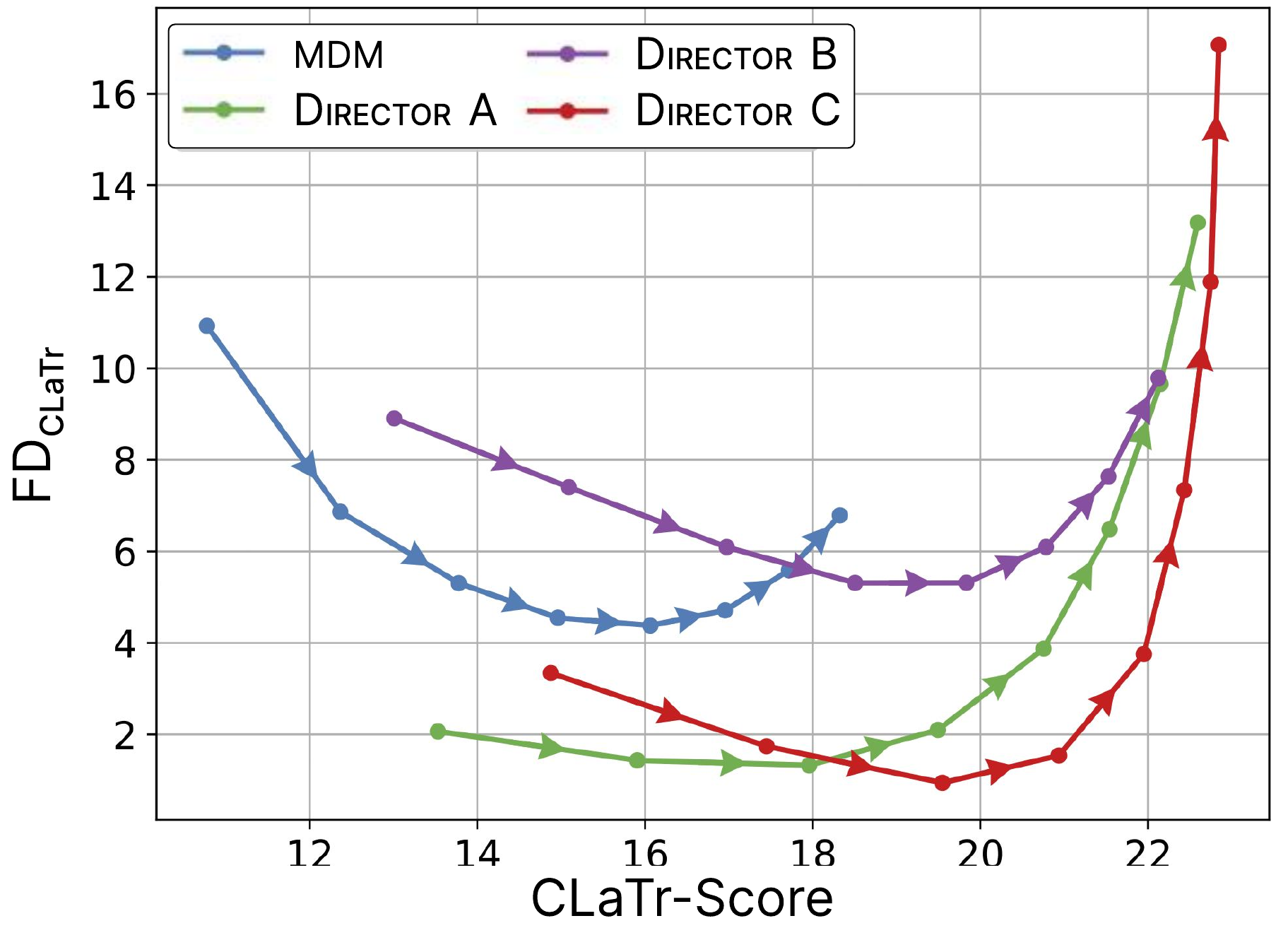}
    \caption{
    \textbf{FD\textsubscript{CLaTr} vs CLaTr-Score.} Guidance range between 0.6 and 2.2 on E.T. mixed subset.}
    \label{fig:fcd-vs-clatr}
  \end{minipage}
\end{figure}

\paragraph{Implementation details.}
We train {\sc Director} with a batch size of 128 using the AdamW optimizer with a learning rate of 1e-4, ($\beta_1,\beta_2) {=} (0.9,0.95)$ and a weight decay of $0.1$. We use a cosine decay learning rate scheduler with 5k steps of warmup for a total of 170k steps in bfloat16 mixed precision. The model has 8 layers with a hidden dim of 512 and 16 attention heads. We use dropout and stochastic depth of 0.1. We set the default temporal input size to 300 to match the E.T. sample size (see Section~\ref{sub:dataset-creation}) and use masking to handle inputs with fewer than 300 frames. For the camera trajectory, we use the 6D continuous representation for rotation~\cite{zhou2019continuity} combined with the 3D translation component. For the character trajectory, we use the 3D position of the character's hip center. 

\subsection{Quantitative results}

\paragraph{Metrics.}
We use two sets of metrics. \\
First, we assess \textbf{camera trajectory quality}, specifically how well the generated camera trajectories match the distribution of the ground truth camera trajectories.
For this, we use the CLaTr-based metrics described in Section~\ref{sec:clatr}: the Frechet CLaTr Distance (FD\textsubscript{CLaTr}) similar to FID~\cite{heusel2017fid}), Precision (R), Recall (R), Density (D) and Coverage (C)~\cite{ferjad2020prdc}. As the validation set comprises only a few samples and these metrics need a critical amount of samples (10k+), we compare to the train set as it is common practice in small dataset generative models (e.g. CIFAR image generation~\cite{krizhevsky2009learning,ho2020ddpm}). \\
Second, we use \textbf{text-camera coherence} metrics, which measure the coherence between the given caption (text) and the generated camera trajectory. For this, we use the CLaTr-Score (CS) (see Section~\ref{sec:clatr}), similar to CLIP-Score~\cite{hessel2021clipscore}. 
Additionally, we derive Classifier Precision (C-P), Classifier Recall (C-R) and Classifier F1-Score (C-F1) by performing motion tagging (described in Section~\ref{sub:dataset-creation}) on generated camera trajectories and compare them to the ground truth.

\paragraph{Dataset.} 
In our experiments, we train and evaluate our model on two different subsets of the E.T. dataset. 
First, the \emph{pure camera trajectory subset}, where we only keep the samples having a single camera motion trajectory (e.g. ``\textit{the camera trucks right}''). 
Second, the \emph{mixed camera trajectories subset}, which excludes some static-only camera trajectories to create a balanced subset. 
In this way, we can both correctly compare against methods suited for simple, pure trajectories and emphasize the difficulty of the mixed compositional camera trajectories. 
We compare in Table~\ref{tab:quant} {\sc Director} with concurrent methods on the pure subset (top) and mixed subset (bottom).

\paragraph{Comparison to the state of the art.}
We report in Table~\ref{tab:quant} and Figure~\ref{fig:fcd-vs-clatr} quantitative results of the different {\sc Director} architectures against the previous state-of-the-art CCD~\cite{jiang2024ccd}, and MDM~\cite{tevet2022mdm}, a default modern method in human motion. 
We observe that overall we outperform both works on all metrics and both subsets. 
Particularly, in the mixed trajectory subset (bottom of Table~\ref{tab:quant}), we demonstrate superior camera trajectory quality metrics (left section of Table~\ref{tab:quant}) with a margin of $-3.0$ FD\textsubscript{CLaTr} against MDM and $-32.1$ against CCD. Additionally, our method excels in text-camera coherence (right section of Table~\ref{tab:quant}) within the same subset, achieving a substantial improvement of $+3.6$ ClaTR-Score against MDM and $+15.7$ against CCD.

Additionally, we show in Figure~\ref{fig:fcd-vs-clatr} the trade-off between FD\textsubscript{CLaTr} (trajectory quality) and CLaTr-Score (conditioning coherence) for varying guidance weights. The optimal point is at the bottom right, where FD\textsubscript{CLaTr} is lowest and CLaTr-Score is highest. 
We observe that the MDM curve (blue) consistently lies above {\sc Director}'s curves, indicating that MDM performs worse.

These results reveal the effectiveness of our method both in generating high-quality camera trajectories and in handling the input caption conditioning.

\paragraph{Ablation of {\scshape Director} architectures.} We observe in Table~\ref{tab:quant} and Figure~\ref{fig:fcd-vs-clatr} that {\scshape Director} C outperforms other variants, followed closely by {\scshape Director} A. 
The cross-attention mechanism in {\scshape Director} C enables effective incorporation of conditioning into the model, leading to its superior performance. 
{\scshape Director} A offers a compelling balance of efficiency and performance: it exhibits comparable results to {\scshape Director} C with a simpler concept and fewer parameters. 
In contrast, {\scshape Director} B excels in text-camera coherence on the pure trajectory subset (top-right of Table~\ref{tab:quant}) but struggles on the mixed trajectory subset (bottom-right of Table~\ref{tab:quant}). 
We attribute this to the AdaLN's ability to condition the model in simple setups, but its failure to capture sequential complexity in harder scenarios.

\subsection{Qualitative results}

\begin{figure}[t]
\centering
\begin{subfigure}[b]{.24\linewidth}
\includegraphics[width=0.95\linewidth]{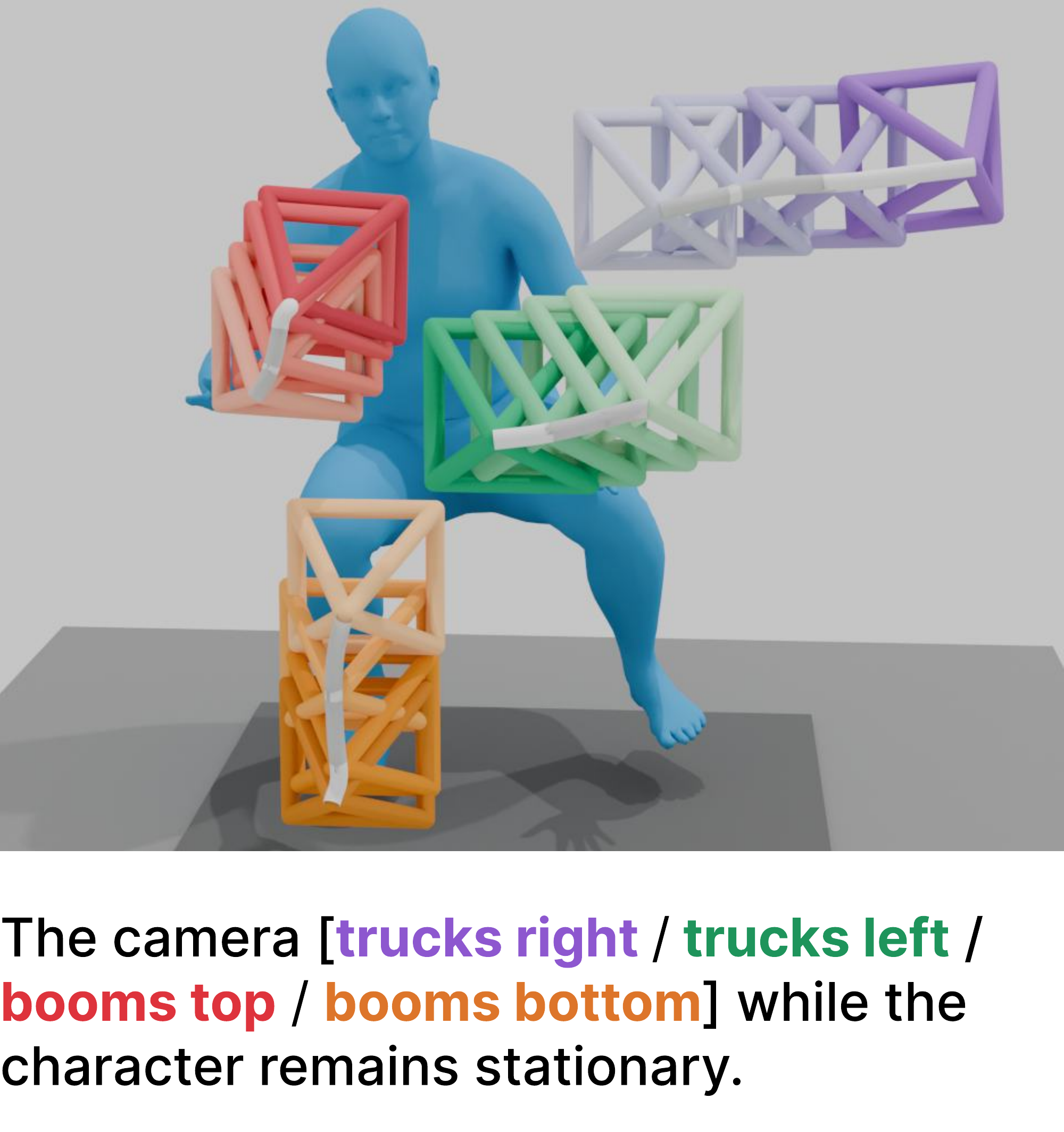}
\caption{Controllability}\label{fig:controllability}
\end{subfigure}
\begin{subfigure}[b]{.24\linewidth}
\includegraphics[width=0.95\linewidth]{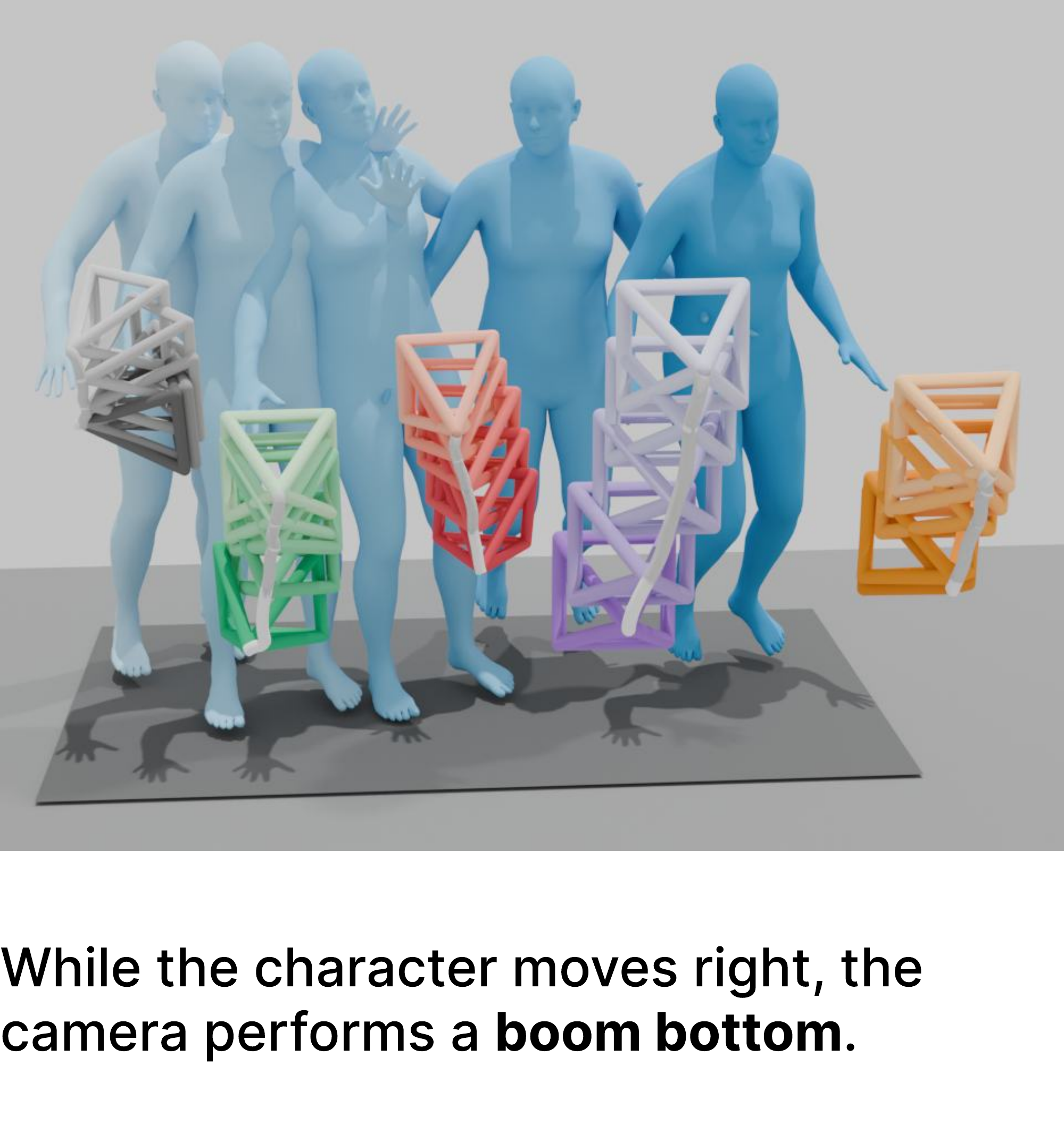}
\caption{Diversity}\label{fig:diversity}
\end{subfigure}
\begin{subfigure}[b]{.24\linewidth}
\includegraphics[width=0.95\linewidth]{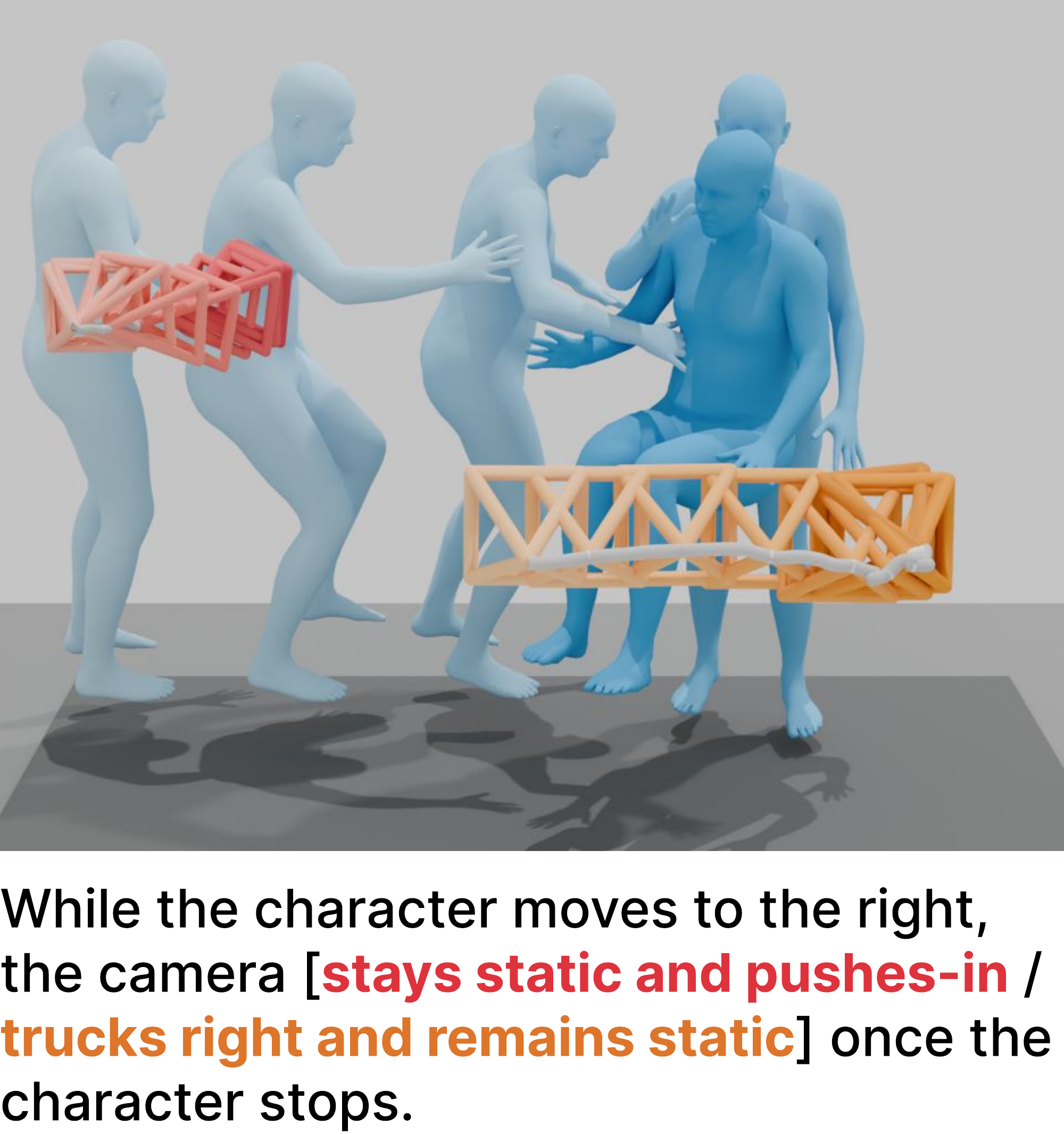}
\caption{Complexity}\label{fig:complexity}
\end{subfigure}
\begin{subfigure}[b]{.24\linewidth}
\includegraphics[width=0.95\linewidth]{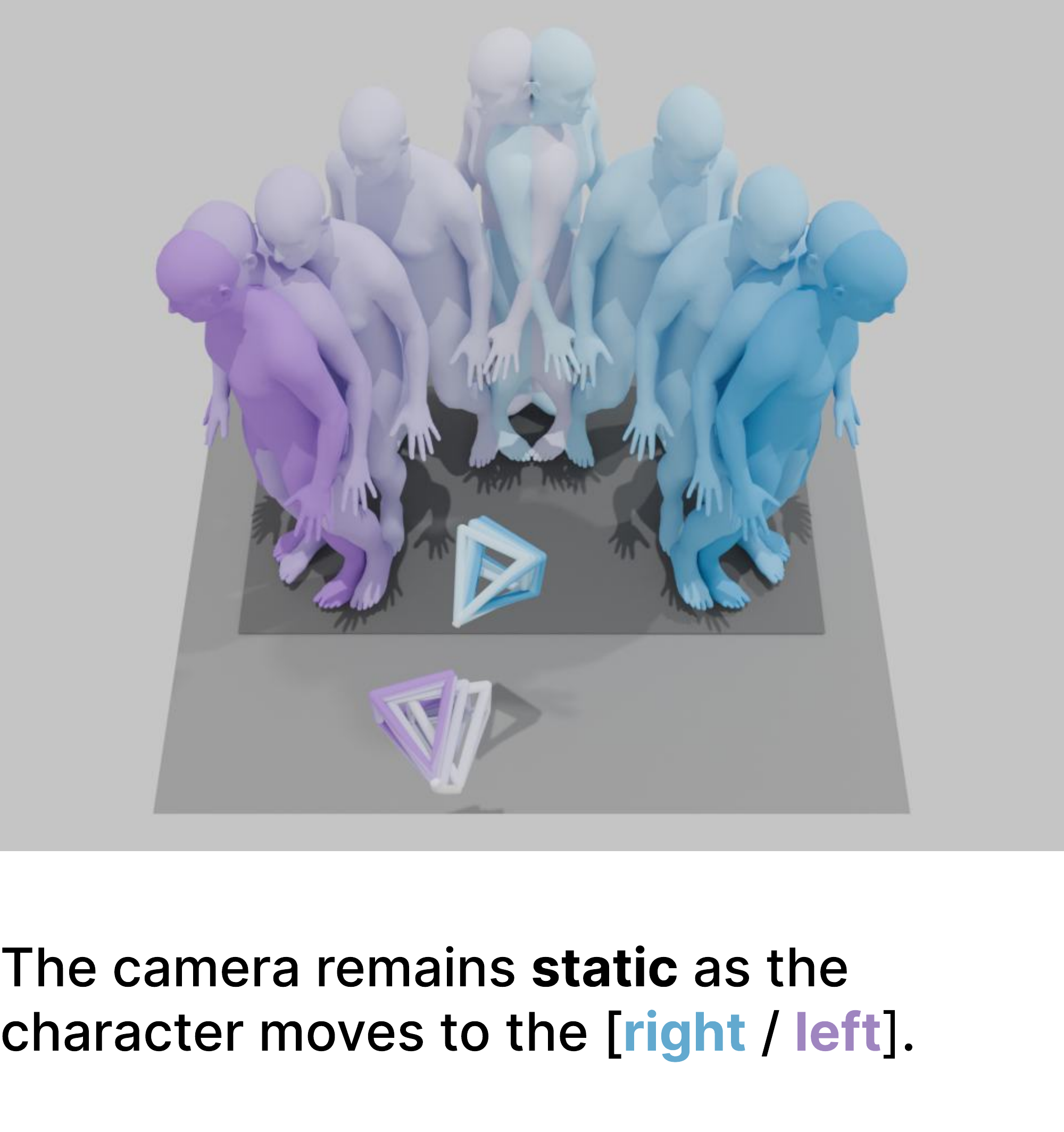}
\caption{Character-aware}\label{fig:character}
\end{subfigure}

\caption{
\textbf{Qualitative results.}
Generated camera trajectories with corresponding prompts and character trajectories, highlighting (a) controllability, (b) diversity, (c) complexity, and (d) character awareness. Darker shades indicate later frames.
}
\label{fig:examples}
\end{figure}

Figure~\ref{fig:examples} shows generated camera trajectories from {\sc Director} (architecture C). Each sub-figure displays the trajectories with pyramid markers for keyframes, along with character meshes and corresponding captions. The output trajectories are smooth and consistent with the input conditions. We highlight four key strengths of our method:

\noindent \textbf{Controllability (Figure~\ref{fig:controllability}).} {\sc Director} offers high controllability: by modifying only two words in the caption, the user can generate all kinds of camera trajectories, e.g. ``\textcolor{custompurple}{\textit{trucks right}}'', ``\textcolor{customgreen}{\textit{trucks left}}'', ``\textcolor{customred}{\textit{booms top}}'' and ``\textcolor{customorange}{\textit{booms bottom}}''.

\noindent \textbf{Diversity (Figure~\ref{fig:diversity}).} Given the same input conditions (i.e. character trajectory and caption), {\sc Director} generates diverse camera trajectories, allowing users to explore a wide range of creative and unique outputs.

\noindent \textbf{Complexity (Figure~\ref{fig:complexity}).} {\sc Director} can handle complex input conditions, including character trajectories (e.g., ``\textit{moves right}'' then ``\textit{stops}'') and camera trajectories descriptions (e.g., ``\textcolor{customred}{\textit{stays static and pushes-in}}'' and ``\textcolor{customorange}{\textit{trucks right and remains static}}'').

\noindent \textbf{Character-awareness (Figure~\ref{fig:character}).} {\sc Director} effectively considers the character, generating camera trajectories that follow the character's movement when the prompt and character trajectory are mirrored.

\section{Conclusion}
\label{sec:conclusion}
We designed and implemented E.T., a dataset of camera and character trajectories extracted from movie sequences that we believe will be very beneficial to the community. In addition to their trajectories, E.T. comes with text captions that describe the camera and character trajectories over time. We showed how E.T. can be exploited to train a diffusion-based approach to generate complex camera trajectories from high-level textual descriptions which correlate the trajectory of the camera with the trajectory of the characters. For this, we propose the diffusion-based method {\sc Director}, which sets the new state of the art on camera trajectory generation. 
In the future, we plan to address the expressiveness of the trajectory captions, by including more information about modifiers and the exact position on the screen where the characters should be located.

\section*{Acknowledgements}
\label{sec:acknowledgements}
This work was supported by ANR-22-CE23-0007, ANR-22-CE39-0016, Hi!Paris grant and fellowship, and was granted access to the HPC resources of IDRIS under the allocation 2023-AD011013951 made by GENCI. 
We would like to thank Hongda Jiang, Mathis Petrovich, Pierre Vassal and the anonymous reviewers for their insightful comments and suggestions.

\newpage

%
%
\bibliographystyle{splncs04}
\bibliography{short_strings,references}

\newpage
\clearpage
\renewcommand{\thesection}{\Alph{section}}
\appendix
\section*{Appendix}
\section{Ethical discussion}

We discuss the ethical impact of our method across several aspects:
\begin{itemize}
    \item \textit{Creative Integrity:} It is a fine line between using AI tool to enhance the human creativity and allowing it to deprive human creative process. Under misusage, the proposed method could diminish the artistic expression instead of support it.
    \item \textit{Intellectual Property}: The use of AI-generated content raises questions about ownership and copyright. The Intellectual Property ownership of the generated content can be debatable.
    \item \textit{Job Displacement or Creation}: The automation of certain aspects of filmmaking could lead to concerns about job displacement within the industry, or under proper usage, may also help to create new types of jobs in the domain.
\end{itemize}

\section{Exceptional Trajectories dataset (E.T.)}

\begin{figure}[htp]
  \centering
    \includegraphics[width=0.85\linewidth]{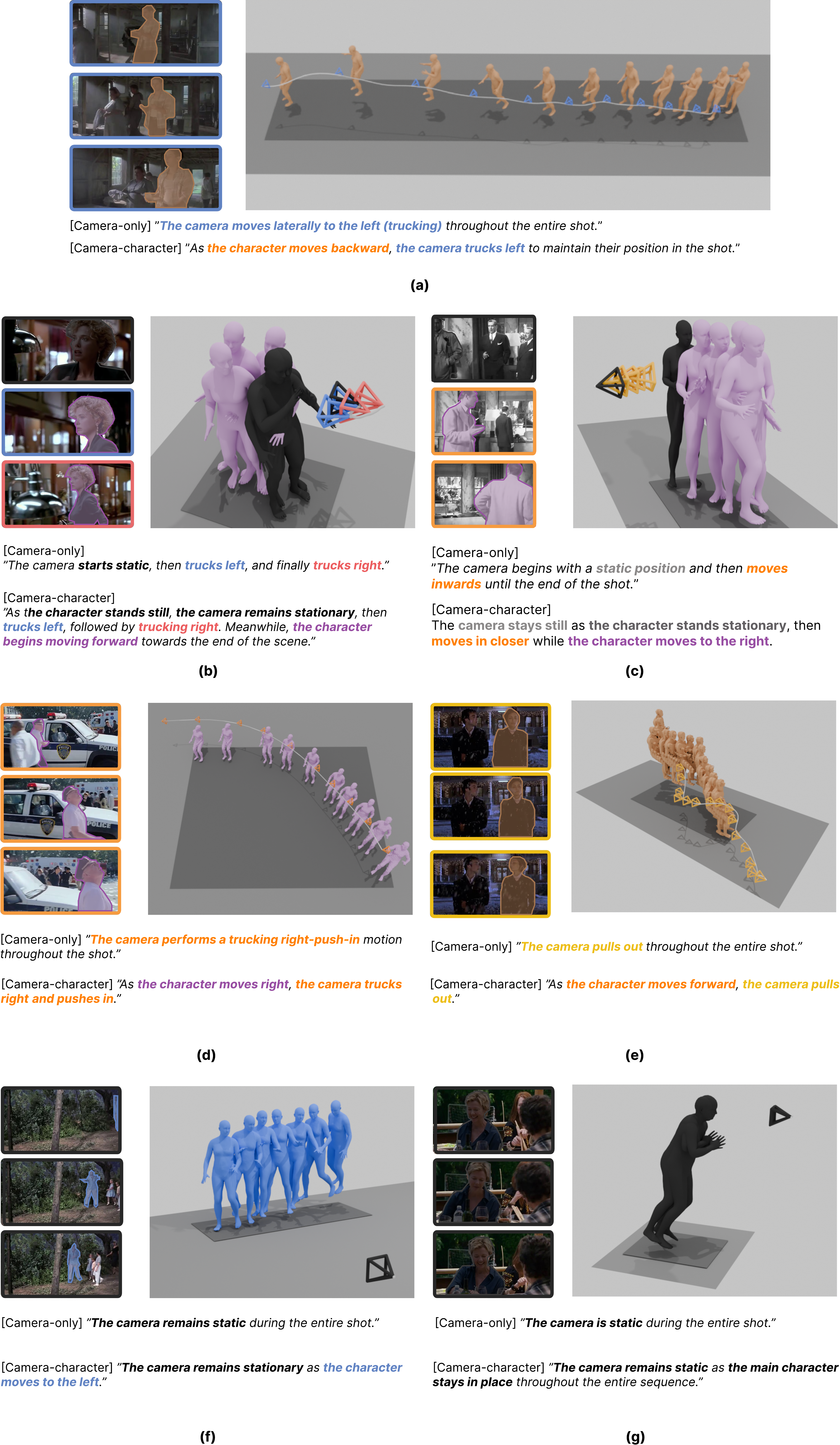}
    \caption{\textbf{Examples E.T. samples.} Each subfigure presents frames from the original movie shot (left), and processed camera and character trajectories (right). Additionally, the bottom part showcases the generated camera trajectory caption with or without the character trajectory caption.}
    \label{sup:fig:example-dataset}
\end{figure}

\subsection{Additional statistics}
\label{supp:sub:stats}

\begin{figure}[t]
\centering

\begin{subfigure}[b]{.3\linewidth}
\scalebox{.23}{\input{fig/cam_frames_hist.pgf}}
\caption{Trajectory length (in \\\#frames)}\label{sup:fig:sample-length}
\end{subfigure}
\begin{subfigure}[b]{.3\linewidth}
\scalebox{.23}{\input{fig/cam_meters_hist.pgf}}
\caption{Camera length (in \\meters)}\label{sup:fig:camera-length}
\end{subfigure}
\begin{subfigure}[b]{.3\linewidth}
\scalebox{.23}{\input{fig/char_meters_hist.pgf}}
\caption{Character length (in \\meters)}\label{sup:fig:character-length}
\end{subfigure}

\caption{\textbf{E.T. statistics.} 
}
\label{sup:fig:dataset-distribution}
\end{figure}

We build our E.T. dataset the Condensed Movies Dataset~\cite{bain2020cmd} (CMD), encompassing over $30,000$ scenes from $3,000$ diverse movies, totaling more than $1,000$ hours of video.
We segment each movie scene into continuous shots by leveraging changes in color and intensity between frames~\cite{pyscenedetect}.

We show additional statistics of E.T. in Figure~\ref{sup:fig:dataset-distribution}. We observe that for both camera and character, the majority of trajectories are smaller than 20 meters, i.e. corresponding to a velocity of $20 \text{ meters} / (300 \text{ frames} / 25 \text{ fps}) = 1.67 m.s^{-1}$.

Additionally, in Figure~\ref{sup:fig:example-dataset}, we show extensive examples of E.T. samples.

\subsection{Data pre-processing}
\label{supp:sub:processing}

\begin{figure}[t]
\centering
\begin{subfigure}[b]{.45\linewidth}
\includegraphics[width=0.95\linewidth]{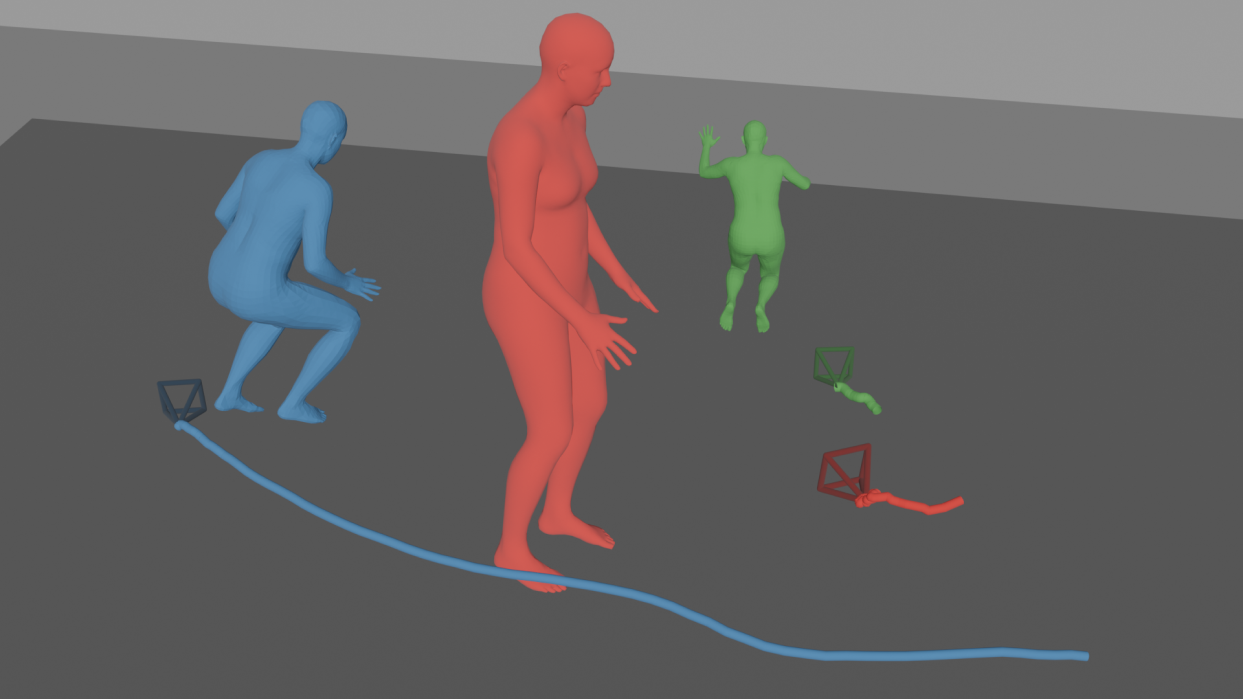}
\caption{Before alignment.}\label{sup:fig:chunk-dataset}
\end{subfigure}
\begin{subfigure}[b]{.45\linewidth}
\includegraphics[width=0.95\linewidth]{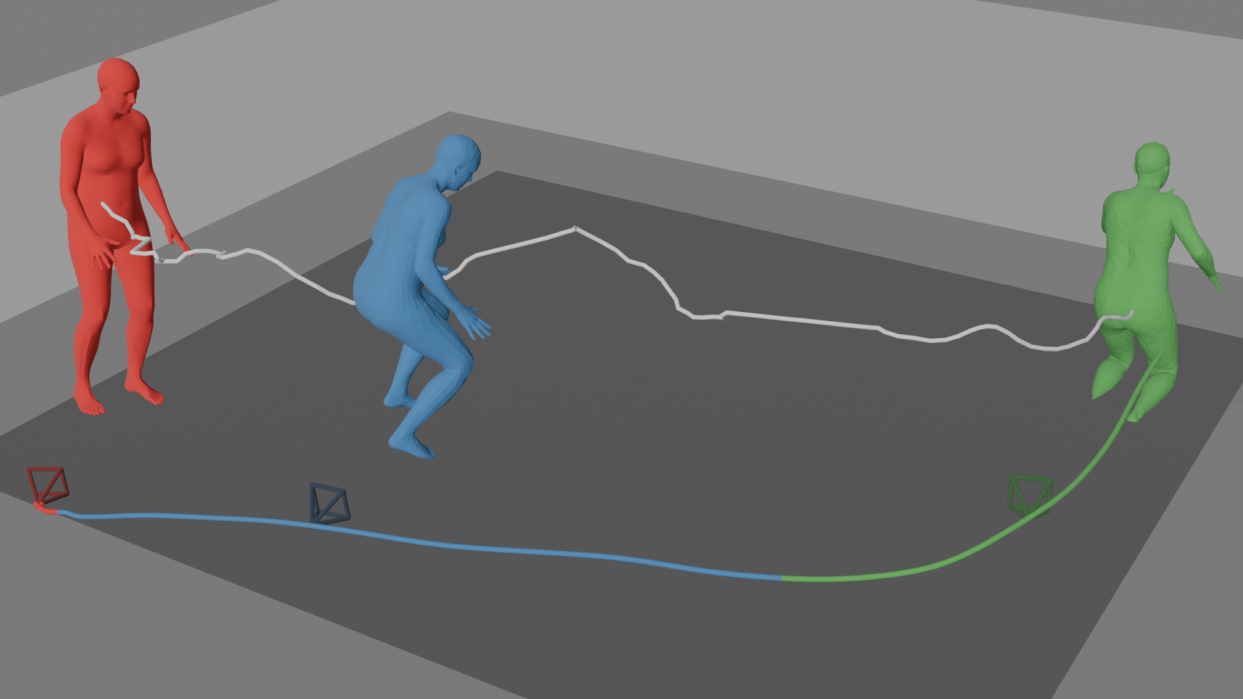}
\caption{After alignment.}\label{sup:fig:align-dataset}
\end{subfigure}
\caption{\textbf{Raw chunk alignment.} We show in (a) the raw independent chunks just after the SLAHMR~\cite{ye2023slahmr} extraction. In (b) we display the result of the chunk alignment process. Each color (red, blue, green) corresponds to a different chunk. 
}

\label{sup:fig:chunks}
\end{figure}

\paragraph{Chunk alignment.}
A limitation of SLAHMR~\cite{ye2023slahmr} is its inability to handle long videos (exceeding 100 frames). Consequently, we divide each shot into chunks of 100 frames and process them independently. However, it produces non-consitant outputs: it exhibits translational bias/offset and different scales, as shown in Figure~\ref{sup:fig:chunk-dataset}.

To address this issue, we propose the following alignment method: dividing shots into overlapping chunks, where consecutive chunks share frames, and performing alignment on these overlapping frames.
A chunk contains camera trajectories with $SE(3)$ poses represented as $[\mathbf{R} |\mathbf{t}]$ (where $\mathbf{R}$ denotes rotation and $\mathbf{t}$ translation), and 3D human poses described by $\mathbf{V}$ (vertices of a 3D mesh).

Given two consecutive chunks at $k$ and $k+1$, we initially align the cameras. The alignment involves determining a scale parameter $s$ and a $SE(3)$ rigid transformation $[\mathbf{B}\; | \; \mathbf{b}]$: 
\begin{align}
    &[\mathbf{R}_{k}\; | \; \mathbf{t}_{k}] = [\mathbf{B}_{k}\; | \; \mathbf{b}_{k}] \,[\mathbf{R}_{k+1}\; | \; s_{k} \, \mathbf{t}_{k+1}], \\
    &[\mathbf{R}_{k}\; | \; \mathbf{t}_{k}] = [\mathbf{B}_{k} \, \mathbf{R}_{k+1}\; | \; s_{k} \, \mathbf{B}_{k} \,\mathbf{t}_{k+1}+\mathbf{b}_{k}],
\end{align}
which simplifies to:
\begin{align}
    &(a) \quad \mathbf{R}_{k} = \mathbf{B}_{k} \, \mathbf{R}_{k+1},  
    \label{eq:a}
    \\
    &(b) \quad \mathbf{t}_{k} = s_{k} \, \mathbf{B}_{k} \,\mathbf{t}_{k+1} + \mathbf{b}_{k}.
    \label{eq:b}
\end{align}

Notably, the rotation estimated by SLAHMR remains consistent across chunks, implying $\mathbf{B}_{k} = \mathbf{I}$, and simplifying Equations~\ref{eq:a}~and~\ref{eq:b} : 
\begin{align}
    &(a) \quad \mathbf{R}_{k} = \mathbf{R}_{k+1}, \\
    &(b) \quad \mathbf{t}_{k} = s_{k} \, \mathbf{t}_{k+1} + \mathbf{b}_{k}.
\end{align}
Subsequently, alignment entails determining the scaling factor $s$ and translational bias $\mathbf{b}$. These parameters can be accurately estimated using the least-square method \cite{bjorck1990leastsquares}, as represented by:
\begin{align}
    \begin{bmatrix}
        \mathbf{t}_{k} & \mathbf{I}
    \end{bmatrix} 
    \begin{bmatrix}
        s_{k} \\
        \mathbf{b_{k}}
    \end{bmatrix} 
    =
    \mathbf{t}_{k+1},
\label{eq:cam-align}
\end{align}
which can be further expressed as:
\begin{align}
\begin{bmatrix}
t_{k}^x & 1 & 0 & 0 \\
t_{k}^y & 0 & 1 & 0 \\
t_{k}^z & 0 & 0 & 1 
\end{bmatrix} 
\begin{bmatrix}
s_{k} \\
b^x_{k} \\
b^y_{k} \\
b^z_{k} 
\end{bmatrix} 
=
\begin{bmatrix}
t_{k+1}^x \\
t_{k+1}^y \\
t_{k+1}^z 
\end{bmatrix} .
\end{align}
We also seek the alignment transform $\Delta_b$, such that:
\begin{align}
[\mathbf{R}_{k+1}\; | \; s_{k} \, \mathbf{t}_{k+1}+\mathbf{b}_{k}] \, \mathbf{\Delta}_b = [\mathbf{R}_{k+1} \, | \, \mathbf{t}_{k+1}],
\end{align}
resulting in:
\begin{equation}
\mathbf{\Delta}_b = [\mathbf{R}_{k+1}\; | \; s_{k} \, \mathbf{t}_{k+1}+\mathbf{b}_{k}]^{-1}\, [\mathbf{R}_{k+1} \, | \, \mathbf{t}_{k+1}].
\label{eq:delta-align}
\end{equation}
Considering the inverse of a 4x4 transformation matrix representing a rigid transformation:
\begin{align}
\begin{bmatrix}
\mathbf{R}^T & -\mathbf{R}^T \mathbf{t} \\
\mathbf{0} & 1
\end{bmatrix} ,
\end{align}
we obtain from Eq.~\ref{eq:delta-align}:
\begin{align}
    & \mathbf{\Delta}_b =
    \begin{bmatrix}
        \mathbf{R}_{k+1}^T & -\mathbf{R}_{k+1}^T (s\mathbf{t}_{k+1}+\mathbf{b}_{k}) \\
        \mathbf{0} & 1
    \end{bmatrix} \,
    \begin{bmatrix}
        \mathbf{R}_{k+1} & \mathbf{t}_{k+1} \\
        \mathbf{0} & 1
    \end{bmatrix}, \\
    & \mathbf{\Delta}_b = 
    \begin{bmatrix}
        \mathbf{I} & \mathbf{R}_{k+1}^T (\mathbf{t}_{k+1} - (s\mathbf{t}_{k+1}+\mathbf{b}_{k})) \\
        \mathbf{0} & 1
    \end{bmatrix}.
\end{align}
Ultimately, to align the 3D human poses based on their vertices $V$:
\begin{align}
    & \begin{bmatrix} 
        \mathbf{V}_{k}^T \\
        1
    \end{bmatrix} = \mathbf{\Delta}_b \, \begin{bmatrix} 
        \mathbf{V}_{k+1}^T \\
        1
    \end{bmatrix} = \begin{bmatrix} 
        \mathbf{V}_{k+1}^T + \mathbf{R}_{k+1}^T (\mathbf{t}_{k+1} - (s_{k}\mathbf{t}_{k+1}+\mathbf{b}_{k})) \\
        1
    \end{bmatrix} ,
\end{align}
\begin{align}
        \mathbf{V}_{k} =  
        \mathbf{V}_{k+1} + (\mathbf{t}_{k+1} - (s_{k}\mathbf{t}_{k+1}+\mathbf{b}_{k}))^T \mathbf{R}_{k+1} .
\end{align}
The alignment process outcome is illustrated in Figure~\ref{sup:fig:align-dataset}.

\paragraph{Data cleaning.}
The extracted trajectories have limitations from the data extraction method~\cite{ye2023slahmr}, including discontinuities, ruptures and jerky motions.
To address this, we first clean the data by removing outliers (i.e., discontinuous segments), with a velocity threshold. Specifically, we eliminate trajectory points holding velocities greater than the 95th percentile of the overall trajectory velocity multiplied by a scaling factor. 
Subsequently, the trajectory is partitioned into sub-trajectories without outliers. 
Finally, we use Kalman filter on each chunk to reduce residual jerkiness and enhance overall smoothness.

\subsection{Dataset creation pipeline}
\label{supp:sub:pipeline}

\paragraph{Motion tagging.} We tune the parameters of our motion tagging method using the dataset introduced in~\cite{courant2021high}. This small dataset of 75 short clips includes annotated sequences of pure camera motion. For the character trajectory tagging, we extended this dataset by annotating human trajectories.
We select parameters (i.e. mainly threshold values) that corresponds to the best classification metrics described in Section~\ref{sec:experiments} of the main manuscript.

\paragraph{Caption generation.}

We show the prompt used for caption generation (see Section~\ref{sub:dataset-creation} of the main manuscript):
\begin{verbatim}
You act as a camera operator writing a technical script for camera 
motion descriptions.

Given a rough outline of the camera motion and main character motion, 
write the camera motion description according to the main character 
motion.

The sentence should be short, and factual. Do not mention frame 
indices.

# Examples
Outline: Total frames 209.
    [Camera motion] Between frames 0 and 154: boom top, Between 
    frames 155 and 209: static.
    [Main character motion] Between frames 0 and 146: move up, 
    Between frames 147 and 209: static.
Description: While the character climbs up, the camera follows them 
with a boom top, and as soon as the character stops, it remains 
static.
# End of examples

Outline: Total frames {CURRENT_NUM_FRAME}.
    [Camera motion] {CURRENT_CAMERA_DESCRIPTION}.
    [Main character motion] {CURRENT_CAMERA_DESCRIPTION}.
Description: 
\end{verbatim}


\section{Contrastive Language-Trajectory embedding (CLaTr)}
\label{sup:sub:clatr}

\begin{figure}[t]
\centering
\begin{subfigure}[b]{.59\linewidth}
\includegraphics[width=0.99\linewidth]{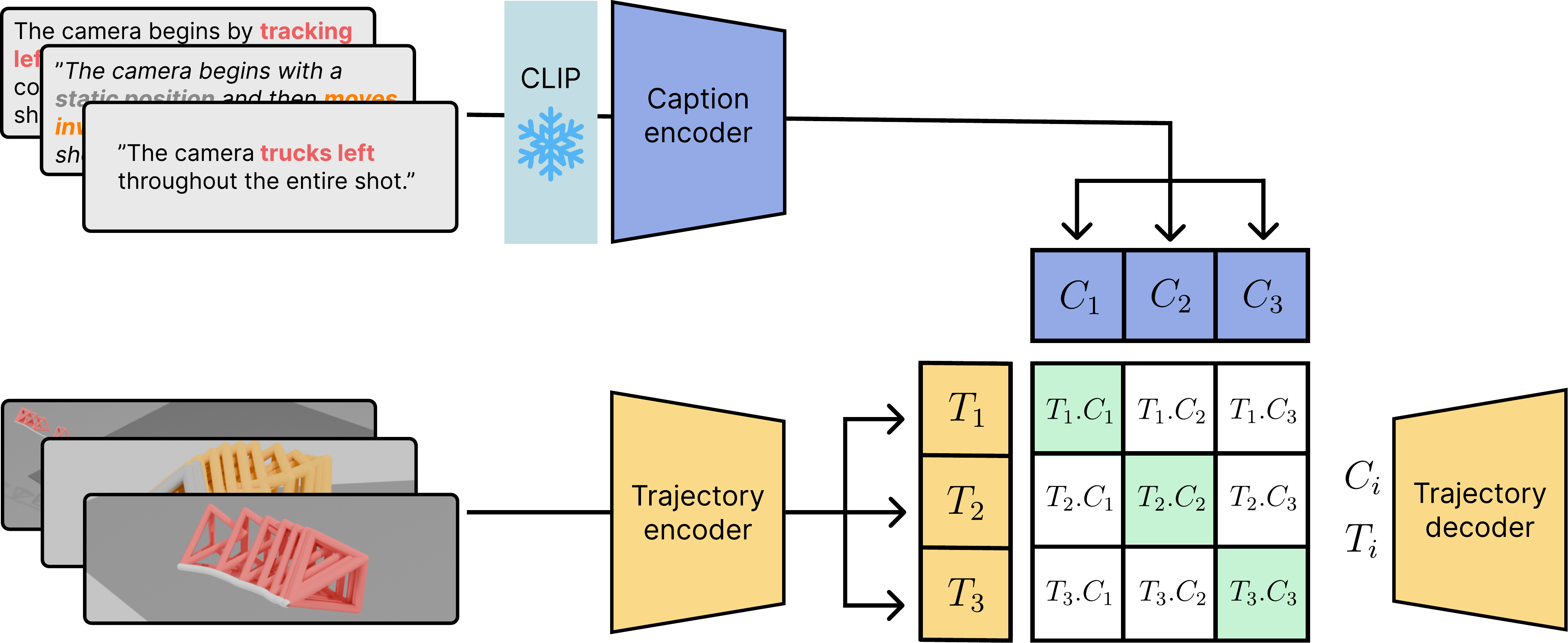}
\vspace{10pt}
\caption{\textbf{Overview of CLaTr framework.} CLaTr projects both text and camera trajectories into a common latent space using encoders. Self-similarity is then computed, and a shared-weight decoder decodes both text and camera trajectory features back into a camera trajectory.}\label{sup:fig:main-clatr}
\end{subfigure}
\begin{subfigure}[b]{.39\linewidth}
\includegraphics[width=0.95\linewidth]{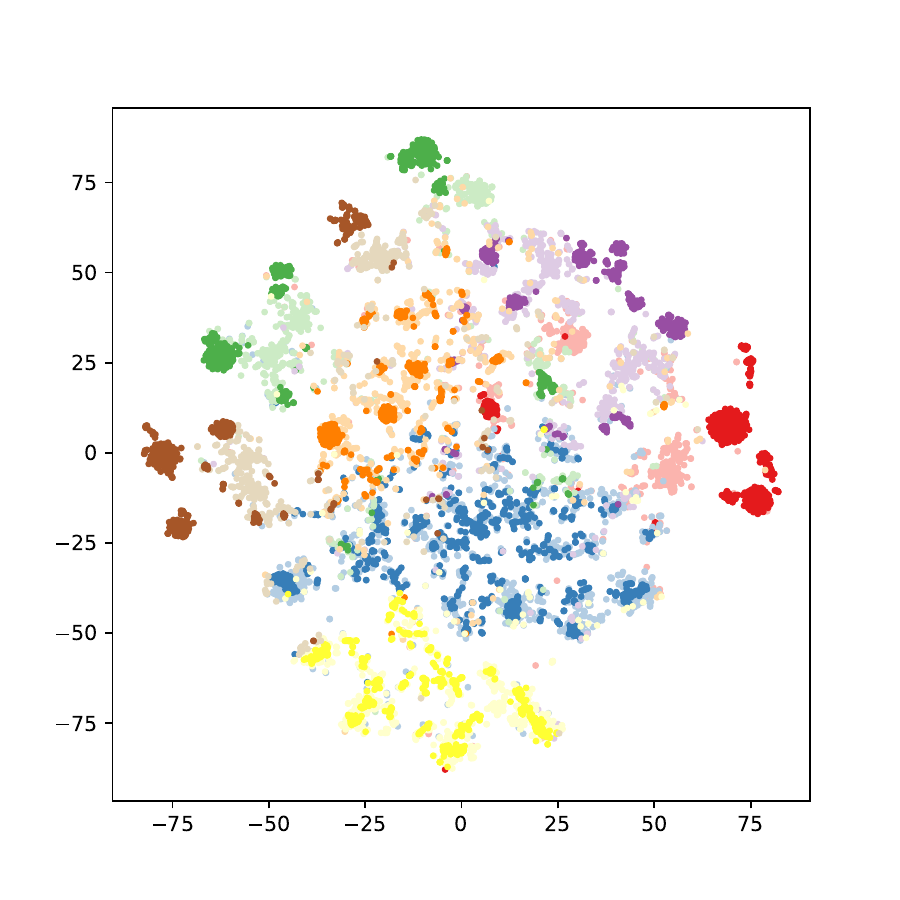}
\caption{\textbf{t-SNE visualization of CLaTr embedding} of text (vivid colors) and trajectory (pastel colors). Each color corresponds to a K-Mean cluster of the text embedding.}\label{sup:fig:tsne-clatr}
\end{subfigure}
\label{sup:fig:clatr}
\end{figure}

\newcolumntype{A}{>{\hspace{6pt}}c<{\hspace{6pt}}}

\begin{table}[htbp]
\centering
\resizebox{\textwidth}{!}{
\begin{tabular}{AAAAAA|AAAAAA}
\toprule
\multicolumn{6}{c|}{\textbf{Text-trajectory retrieval}} & \multicolumn{6}{c}{\textbf{Trajectory-text retrieval}} \\
R@1 $\uparrow$  & R@2 $\uparrow$  & R@3 $\uparrow$  & R@5 $\uparrow$  & R@10 $\uparrow$ & MedR $\downarrow$ & R@1 $\uparrow$  & R@2 $\uparrow$  & R@3 $\uparrow$  & R@5 $\uparrow$  & R@10$\uparrow$ & MedR $\downarrow$ \\
\midrule
$19.73$     &  $31.67$     &   $40.8$    &  $52.08$    &    $64.69$   &   $5.0$    & $11.15$ &   $17.25$    &    $20.91$   &    $26.5$   &     $34.66$  &   $28.0$  \\
\bottomrule
\end{tabular}
}
\caption{\textbf{CLaTr evaluation.} We report the retrieval scores of CLaTr on the E.T. dataset.}
\label{tab:clatr}
\end{table}

We show in Figure~\ref{sup:fig:main-clatr} the overview of the CLaTr framework as described in Section~\ref{sec:clatr} of the main manuscript.

\paragraph{Implementation details.} We train CLaTr with a batch size of $32$ using the AdamW optimizer with a learning rate of $1e-5$.  
The set the weight of the reconstruction loss at $1.0$, of the latent loss at $1.0e-5$, of the KL loss at $1.0e-5$, and of the contrastive loss at $0.1$. The model has $6$ layers with a hidden dim of $256$ and $4$ attention heads. We use dropout of $0.1$.
Similar to {\sc Director}, we set the default temporal input size to 300 and use masking to handle inputs with fewer than 300 frames. We represent the camera trajectory with the 6D continuous representation for rotation~\cite{zhou2019continuity} combined with the 3D translation component.

\paragraph{CLaTr Evaluation.} Table~\ref{tab:clatr} presents standard retrieval performance measures from~\cite{petrovich2023tmr,guo2022humanml3d}. Recall at rank k (R@k) indicates the percentage of times the correct caption is within the top k results (higher is better). Median rank (MedR) is also reported, where lower values are better.

As shown in Table~\ref{tab:clatr}, text-to-trajectory metrics outperform trajectory-to-text metrics. This may be because text descriptions are more ambiguous and varied in describing trajectories, making it easier to match a text description to a unique trajectory than to match a trajectory to a specific description among many possibilities.

\paragraph{CLaTr embedding.} We show in Figure~\ref{sup:fig:tsne-clatr} a t-SNE visualization of CLaTr text (vivid colors) and trajectory (pastel colors) embeddings. 
We applied K-Means clustering to the text embeddings and visualized the corresponding clusters on the trajectory embeddings to assess the consistency of the joint embedding. 
Notably, we find that text clusters are preserved in the trajectory space, with vivid and pastel clusters overlapping, indicating a robust alignment between text and trajectory representations.

\end{document}